\documentclass[onecolumn]{article}
\usepackage{cite}
\usepackage{natbib}
\usepackage{times}

\usepackage{epsfig}
\usepackage{graphicx}
\usepackage{amsmath}
\usepackage{amssymb}
\usepackage{dsfont}
\usepackage{subfigure}
\usepackage[small]{caption}
\usepackage[colorlinks,citecolor=black]{hyperref}
\usepackage{xspace}
\usepackage{bm}
\usepackage{color}
\newcommand{\vade}{VaDE\xspace}

\newcommand{\reals}{\mathbb{R}}
\newcommand{\realsp}{\mathbb{R}_+}

\allowdisplaybreaks

\title{Variational Deep Embedding: \\An Unsupervised and Generative Approach to Clustering\footnote{This paper
is accepted by IJCAI 2017, http://ijcai-17.org/accepted-papers.html}}

\author{
Zhuxi Jiang$^1$, Yin Zheng$^2$, Huachun Tan$^1$, Bangsheng Tang$^3$, Hanning Zhou$^3$\\
$^1$Beijing Institute of Technology, Beijing, China\\$^2$Tencent AI Lab, Shenzhen, China\\$^3$Hulu LLC., Beijing, China\\
{\tt\small\{zjiang, tanhc\}@bit.edu.cn, yinzheng@tencent.com,}\\
{\tt\small bangsheng.tang@gmail.com, eric.zhou@hulu.com}
}

\begin{document}

\maketitle
\thispagestyle{empty}
\pagestyle{empty} 

\begin{abstract}
Clustering is among the most fundamental tasks in machine learning and artificial intelligence. 
In this paper, we propose Variational Deep Embedding (\vade), a novel unsupervised generative
clustering approach within the framework of Variational Auto-Encoder (VAE). Specifically, \vade models the data generative procedure with a Gaussian Mixture 
Model (GMM) and a deep neural network (DNN): 1) the GMM picks a cluster; 2) from which a latent embedding
is generated; 3) then the DNN decodes the latent embedding into an observable. Inference in \vade is done in a variational way: a different DNN is used to encode observables to latent embeddings, so that the evidence lower bound (ELBO) can be optimized using the Stochastic Gradient Variational Bayes (SGVB) estimator and the {\it reparameterization} trick. Quantitative comparisons with strong baselines are included in 
this paper, and experimental results show that \vade 
significantly outperforms the state-of-the-art clustering methods on $5$ benchmarks from various
modalities. Moreover, by \vade's generative nature, we show its capability of generating highly
realistic samples for any specified cluster, without using supervised information 
during training. 
\end{abstract}

\section{Introduction}
\label{Se1}

\begin{figure}[ht]
\begin{center}
\includegraphics[width=0.8\linewidth]{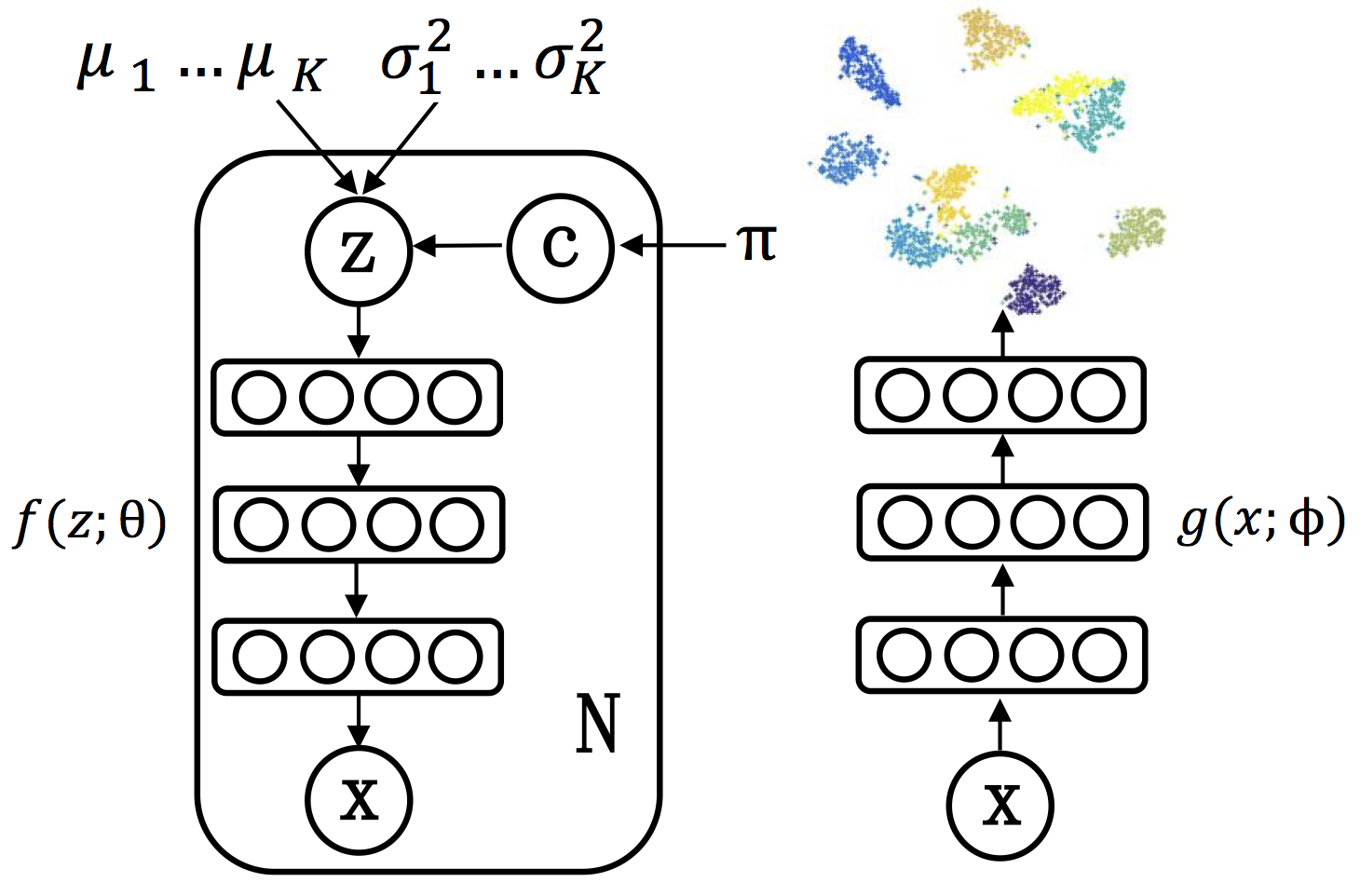}
\end{center}
   \caption{ The diagram of \vade. The data generative process of \vade is done as follows:
   1) a cluster is picked from a GMM model; 2) a latent embedding is generated based on the picked cluster;
   3) DNN $f({\bf z};{\boldsymbol{\theta}})$ decodes the latent embedding into an observable $\bf x$. A encoder network $g({\bf x};{\boldsymbol{\phi}})$
   is used to maximize the ELBO of \vade.
   }
\label{fig:diagram}
\end{figure}

Clustering is the process of grouping similar objects together, which is one of the most fundamental tasks in
machine learning and artificial intelligence. Over the past decades, a large family of clustering algorithms have been developed and successfully applied in enormous real world tasks~\cite{ng02,ye08,yang10,xie15}. 
Generally speaking, there is a dichotomy of clustering methods: Similarity-based clustering and Feature-based clustering.
Similarity-based clustering builds models upon a distance matrix, which is a $N\times N$ matrix that measures the 
distance between each pair of the $N$ samples. One of the most famous similarity-based clustering methods is Spectral Clustering
(SC)~\cite{von07}, which leverages the Laplacian spectra of the distance matrix to reduce dimensionality before clustering. Similarity-based clustering methods have the advantage that domain-specific similarity or kernel functions 
can be easily incorporated into the models. But these methods suffer scalability issue due to super-quadratic running time for computing spectra.

Different from similarity-based methods, a feature-based method takes a $N\times D$ matrix as input, where $N$ is
the number of samples and $D$ is the feature dimension. One popular feature-based clustering method is $K$-means, which aims to
partition the samples into $K$ clusters so as to minimize the within-cluster sum of squared errors. Another representative feature-based
clustering model is Gaussian Mixture Model (GMM), which assumes that the data points are generated from a Mixture-of-Gaussians (MoG), 
and the parameters of GMM are optimized by the Expectation Maximization (EM) algorithm. One advantage of GMM over $K$-means 
is that a GMM can generate samples by estimation of data density. Although $K$-means, GMM and their variants~\cite{ye08,liu10} have been extensively used, learning good representations most suitable for clustering tasks is left largely unexplored. 

Recently, deep learning has achieved widespread success in numerous machine learning 
tasks~\cite{alex12,zheng14sup,szegedy2015going,zheng2014neural,he16,zheng15deep,zheng2016neural}, where learning good representations by deep
neural networks (DNN) lies in the core. Taking a similar approach, it is conceivable to 
conduct clustering analysis on good representations, instead of raw data points. 
In a recent work, Deep Embedded Clustering (DEC)~\cite{xie15} was proposed to 
simultaneously learn feature representations and cluster assignments by deep neural 
networks. Although DEC performs well in clustering, similar to $K$-means, DEC cannot 
model the generative process of data, hence is not able to
generate samples. Some recent works, e.g. VAE~\cite{kingma13}, GAN~\cite{goodfellow14} 
, PixelRNN~\cite{oord2016pixel}, InfoGAN~\cite{Chen16InfoGAN} and PPGN~\cite{nguyen16PPGN},
have shown that neural networks can be trained to generate meaningful samples. 
The motivation of this work is to develop a 
{\it clustering} model based on neural networks that 1) learns good representations that capture the 
statistical structure of the data, and 2)
is capable of generating samples.

In this paper, we propose a clustering framework, Variational Deep Embedding (\vade), that combines 
VAE~\cite{kingma13} and a Gaussian Mixture Model for clustering tasks.
\vade models the data generative process by a GMM and a DNN $f$: 1) a cluster is picked up 
by the GMM; 2) from which a latent representation $\bf z$ is sampled; 3) DNN $f$ decodes $\bf z$ 
to an observation $\bf x$.
Moreover, \vade is optimized by using another DNN $g$ 
to encode observed data $\bf x$ into latent embedding $\bf z$, so that the Stochastic 
Gradient Variational Bayes (SGVB) estimator and the {\it reparameterization} 
trick~\cite{kingma13} can be used to maximize the evidence lower bound (ELBO). 
\vade generalizes VAE in that a Mixture-of-Gaussians prior replaces the single
Gaussian prior. Hence, \vade is by design more suitable for clustering tasks\footnote{Although people 
can use \vade to do unsupervised feature learning or semi-supervised learning
tasks, we only focus on clustering tasks in this work.}. Specifically,
the main contributions of the paper are:

\begin{itemize}
\item We propose an unsupervised generative clustering framework, \vade, that combines VAE and GMM together.
\item We show how to optimize \vade by maximizing the ELBO using the SGVB estimator and the {\it reparameterization} trick;
\item Experimental results show that \vade outperforms the state-of-the-art clustering models on $5$ datasets from various modalities by a large margin;
\item We show that \vade can generate highly realistic samples for any specified cluster, without using supervised information during training.
\end{itemize}
The diagram of \vade is illustrated in Figure~\ref{fig:diagram}.

\section{Related Work}
\label{sec:related_work}

Recently, people find that learning good representations plays an important role in clustering tasks. For 
example, DEC~\cite{xie15} was proposed to learn feature representations and cluster
assignments simultaneously by deep neural networks. In fact, DEC learns a mapping from 
the observed space to a lower-dimensional latent space, where it iteratively optimizes the KL divergence 
to minimize the within-cluster distance of each cluster. DEC achieved impressive performances
on clustering tasks. However, the feature embedding in DEC is designed specifically for clustering
and fails to 
uncover the real underlying structure
of the data, which makes the model lack of the ability to extend itself to other 
tasks beyond clustering, such as generating samples.

The deep generative models have recently attracted much attention in
that they can capture the data distribution by neural networks, 
from which unseen samples can be generated. GAN and VAE are among the most successful deep generative models in recent years.
Both of them are appealing unsupervised generative
models, and their variants have been extensively studied and applied in various tasks such as semi-supervised
classification~\cite{Kingma14Semi,maaloe16auxiliary,salimans16improvedGAN,makhzani16AAE,abbasnejad16infiniteVAE}, 
clustering~\cite{makhzani16AAE} and image generation~\cite{radford15,dosovitskiy16DeePSiM}.

For example, \cite{abbasnejad16infiniteVAE} proposed to use a mixture of VAEs
for semi-supervised classification tasks, where the mixing coefficients of these VAEs are modeled by a 
Dirichlet process to adapt its capacity to the input data. 
SB-VAE~\cite{Nalisnick16SBVAE} also applied Bayesian nonparametric techniques
on VAE, which derived a stochastic latent dimensionality 
by a stick-breaking prior and achieved good performance on semi-supervised classification tasks.
\vade differs with SB-VAE in that the cluster assignment
and the latent representation are jointly considered in the Gaussian mixture prior,
whereas SB-VAE separately models 
the latent representation and the class variable, which fails to capture 
the dependence between them. Additionally, \vade does not need the class label during training,
while the labels of data are required by SB-VAE due to its semi-supervised setting.
Among the variants of VAE, Adversarial Auto-Encoder(AAE)~\cite{makhzani16AAE} can also do 
unsupervised clustering tasks. Different
from \vade, AAE uses GAN to match the aggregated posterior with the prior of VAE, 
which is much more complex than \vade on the training procedure. 
We will compare AAE with \vade in the experiments part.

Similar to \vade, \cite{nalisnickapproximate} proposed DLGMM to combine VAE and GMM together. 
The crucial difference, however, is that \vade uses a mixture of Gaussian prior to 
replace the single Gaussian prior of VAE, which is suitable for clustering tasks
by nature, while DLGMM uses a mixture of Gaussian distribution 
as the approximate posterior of VAE and does not model the class variable.
Hence, \vade generalizes VAE to clustering tasks, whereas DLGMM is used to improve the capacity of the original VAE and is not suitable for
clustering tasks by design. The recently proposed GM-CVAE~\cite{shu16stochastic} 
also combines VAE with GMM together. However, the GMM in GM-CVAE is used to model the transitions between video
frames, which is the main difference with \vade.

\section{Variational Deep Embedding}
\label{sec:model}

In this section, we describe Variational Deep Embedding (\vade), a model for probabilistic 
clustering problem within the framework of Variational Auto-Encoder (VAE). 

\subsection{The Generative Process}
\label{sec:gen_process}
Since \vade is a kind of unsupervised generative approach to clustering, we herein first
describe the generative process of \vade. Specifically, suppose there are $K$ clusters, 
an observed sample $\mathbf{x}\in \reals^D$ is generated by the following process:
\begin{enumerate}
\item Choose a cluster $c \sim \textup{Cat}(\boldsymbol{\pi})$
\item Choose a latent vector ${\bf z} \sim \mathcal{N}\left(\boldsymbol{\mu}_c,\boldsymbol{\sigma}_c^2{\bf I}\right)$   
\item Choose a sample $\bf x$:
    \begin{enumerate}
    \item If $\bf x$ is binary
    \begin{enumerate}
    \item Compute the expectation vector $\boldsymbol{\mu}_x$
    \begin{equation}
        \boldsymbol{\mu}_x = f({\bf z};\boldsymbol{\theta})
        \label{eqn:f1}
    \end{equation}
    \item Choose a sample ${\bf x} \sim \textup{Ber}(\boldsymbol{\mu}_x)$
    \end{enumerate}
    \item If $\bf x$ is real-valued
    \begin{enumerate}
    \item Compute $\boldsymbol{\mu}_x$ and $\boldsymbol{\sigma}_x^2$
    \begin{equation}
        [\boldsymbol{\mu}_x;\log \boldsymbol{\sigma}_x^2] = f({\bf z};\boldsymbol{\theta})
        \label{eqn:f2}
    \end{equation}
    \item Choose a sample ${\bf x}\sim \mathcal{N}\left(\boldsymbol{\mu}_x,\boldsymbol{\sigma}_x^2\mathbf{I}\right)$
    \end{enumerate}
    \end{enumerate}
\end{enumerate}
where $K$ is a predefined parameter, $\pi_k$ is the prior probability for cluster $k$,
$\boldsymbol{\pi}\in \realsp^K$, $1=\sum_{k=1}^K \pi_k$, $\textup{Cat}(\boldsymbol{\pi})$ 
is the categorical distribution parametrized by $\boldsymbol{\pi}$, $\boldsymbol{\mu}_c$ and 
$\boldsymbol{\sigma}_c^2$ are the mean and the variance of the Gaussian distribution corresponding to cluster $c$, $\bf I$ is an identity matrix, $f({\bf z};\boldsymbol{\theta})$ is a neural 
network whose input is $\bf z$ and is parametrized by $\boldsymbol{\theta}$, $\textup{Ber}(\boldsymbol{\mu}_x)$ 
and $\mathcal{N}(\boldsymbol{\mu}_x,\boldsymbol{\sigma}_x^2)$ are multivariate Bernoulli 
distribution and Gaussian distribution parametrized by $\boldsymbol{\mu}_x$ and $\boldsymbol{\mu}_x,\boldsymbol{\sigma}_x$, 
respectively. The generative process is depicted in Figure~\ref{fig:diagram}.

According to the generative process above, the joint probability $p({\bf x}, {\bf z}, c)$ can be
factorized as:
\begin{equation}
p({\bf x}, {\bf z}, c) = p({\bf x}|{\bf z})p({\bf z}|c)p(c),
\label{eqn:fact_p}
\end{equation}
since $\bf x$ and $c$ are independent conditioned on $\bf z$. And the probabilities are defined as:
\begin{eqnarray}
p(c) &=& \textup{Cat}(c|{\boldsymbol{\pi}})\label{eqn:p_c}\\
p({\bf z}|c) &=&  \mathcal{N}\left({\bf z}|\boldsymbol{\mu}_c,\boldsymbol{\sigma}_c^2{\bf I}\right)\label{eqn:p_zc}\\
p({\bf x}| {\bf z}) &=& \textup{Ber}({\bf x}| {\bf \boldsymbol{\mu}}_x) \quad or\quad \mathcal{N}({\bf x}|\boldsymbol{\mu}_x,\boldsymbol{\sigma}_x^2\mathbf{I})\label{eqn:p_xz}
\end{eqnarray}

\subsection{Variational Lower Bound}
\label{sec:vlowerbound}
A \vade instance is tuned to maximize the likelihood of the given data points. Given 
the generative process in Section~\ref{sec:gen_process},  by using Jensen's inequality, 
the log-likelihood of \vade can be written as:
\begin{flalign}
\log p({\bf x})&=\log\int_{\bf z}\sum_{c}p({\bf x,z},c)d{\bf z}\nonumber\\
&\geq E_{q({\bf z},c|{\bf x})}[\log\frac{p({\bf x,z},c)}{q({\bf z},c|{\bf x})}]=\mathcal{L}_{\textup{ELBO}}({\bf x})\label{eqn:loglikelihood}
\end{flalign}
where $\mathcal{L}_{\textup{ELBO}}$ is the evidence lower bound (ELBO), $q({\bf z},c|{\bf x})$ is the variational 
posterior to approximate the true posterior $p({\bf z},c|{\bf x})$. In \vade, we 
assume $q({\bf z},c|{\bf x})$ to be a mean-field distribution and can be factorized as:
\begin{equation}
q({\bf z},c| {\bf x}) = q({\bf z|x})q(c|{\bf x}).
\label{eqn:va_q}
\end{equation}

Then, according to Equation~\ref{eqn:fact_p} and \ref{eqn:va_q}, the $\mathcal{L}_{\textup{ELBO}}({\bf x})$ 
in Equation~\ref{eqn:loglikelihood} can be rewritten as:
\begin{eqnarray}
\mathcal{L}_{\textup{ELBO}}({\bf x})&=&E_{q({\bf z},c|{\bf x})}\left[\log \frac{p({\bf x,z},c)}{q({\bf z},c|{\bf x})}\right]\nonumber\\
&=&E_{q({\bf z},c|{\bf x})}\left[\log p({\bf x,z},c)-\log q({\bf z},c|{\bf x})\right]\nonumber\\
&=&E_{q({\bf z},c|{\bf x})}[\log p({\bf x}|{\bf z})+\log p({\bf z}|c)\label{eqn:elbo_fact}\\
&\quad&+ \log p(c) -\log q({\bf z}| {\bf x}) - \log q(c|{\bf x})]\nonumber
\end{eqnarray} 

In \vade, similar to VAE, we use a neural network $g$ to model $q({\bf z|x})$:
\begin{eqnarray}
[\boldsymbol{\tilde\mu};\log \boldsymbol{\tilde\sigma}^2]&=&g({\bf x};\boldsymbol{\phi})\label{eqn:g_mu_sigma}\\
q({\bf z}|{\bf x})&=&\mathcal{N}({\bf z};\boldsymbol{\tilde\mu},{\boldsymbol{\tilde\sigma}}^2{\bf I})
\label{eqn:q_z_x}
\end{eqnarray}
where $\boldsymbol{\phi}$ is the parameter of network $g$.

By substituting the terms in Equation~\ref{eqn:elbo_fact} with 
Equations~\ref{eqn:p_c}, \ref{eqn:p_zc}, \ref{eqn:p_xz} and \ref{eqn:q_z_x},
and using the SGVB estimator and the {\it reparameterization} trick,
the $\mathcal{L}_{\textup{ELBO}}({\bf x})$ can be rewritten as:
\footnote{This is the case when the observation $\bf x$ is binary. For the real-valued situation, the ELBO
can be obtained in a similar way.}
\begin{small}
\begin{flalign}
\mathcal{L}_{\textup{ELBO}}({\bf x})
=&\frac{1}{L}\sum_{l=1}^L\sum_{i=1}^D{x_i}\log\boldsymbol{{\mu}}^{(l)}_x|_i+(1-x_i)\log(1-\boldsymbol{{\mu}}^{(l)}_x|_i)\nonumber\\
&-\frac{1}{2}\sum_{c=1}^K\gamma_{c}\sum_{j=1}^J(\log\boldsymbol{\sigma}^2_{c}|_{j}+
\frac{\tilde{\boldsymbol{\sigma}}^2|_j}{\boldsymbol{\sigma}^2_{c}|_{j}}+
\frac{(\tilde{\boldsymbol{\mu}}|_j-\boldsymbol{\mu}_{c}|_{j})^2}{\boldsymbol{\sigma}^2_{c}|_{j}})\nonumber\\
&+\sum_{c=1}^K\gamma_{c}\log \frac{\pi_c}{\gamma_{c}}
+\frac{1}{2}\sum_{j=1}^J(1+\log\tilde{\boldsymbol{\sigma}}^2|_j)\label{eqn:ELBO_detail}
\end{flalign}
\end{small}
where $L$ is the number of Monte Carlo samples in the SGVB estimator,
$D$ is the dimensionality of ${\bf x}$ and $\boldsymbol{\mu}_x^{(l)}$, $x_i$
is the $i$\textsuperscript{th} element of $\bf x$,
$J$ is the dimensionality of $\boldsymbol{\mu}_c$, $\boldsymbol{\sigma}_c^2$, 
$\tilde{\boldsymbol{\mu}}$ and $\tilde{\boldsymbol{\sigma}}^2$, 
and ${\bf \ast}|_j$ denotes the $j$\textsuperscript{th} element of $\bf \ast$,
$K$ is the number of clusters, $\pi_c$ is the prior probability of cluster $c$,
and $\gamma_c$ denotes $q(c|{\bf x})$ for simplicity.

In Equation~\ref{eqn:ELBO_detail}, 
we compute $\boldsymbol{\mu}_x^{(l)}$ as
\begin{equation}
    \boldsymbol{\mu}_x^{(l)}=f({\bf z}^{(l)};{\bf \theta}),
\end{equation}
where ${\bf z}^{(l)}$ is
the $l$\textsuperscript{th} sample from $q({\bf z}|{\bf x})$ by Equation~\ref{eqn:q_z_x} to produce
the Monte Carlo samples. According to the {\it reparameterization} trick, ${\bf z}^{(l)}$ is 
obtained by 
\begin{equation}
    {\bf z}^{(l)}=\boldsymbol{\tilde\mu}+\boldsymbol{\tilde\sigma}\circ\boldsymbol{\epsilon}^{(l)},
\end{equation}
where $\boldsymbol{\epsilon}^{(l)}\sim \mathcal{N}(0 ,{\bf I})$, $\circ$ is element-wise multiplication, and 
$\boldsymbol{\tilde\mu}$, $\boldsymbol{\tilde\sigma}$ are derived by Equation~\ref{eqn:g_mu_sigma}.

We now describe how to formulate $\gamma_c \triangleq q(c|{\bf x})$ in Equation~\ref{eqn:ELBO_detail} to
maximize the ELBO. Specifically, $\mathcal{L}_{\textup{ELBO}}({\bf x})$
can be rewritten as:
\begin{small}
\begin{flalign}
&\mathcal{L}_{\textup{ELBO}}({\bf x})=E_{q({\bf z},c|{\bf x})}\left[\log \frac{p({\bf x,z},c)}{q({\bf z},c|{\bf x})}\right]\nonumber\\
=&\int_{\bf z}\sum_c q(c| {\bf x})q({\bf z|x})\left[\log\frac{p({\bf x|z})p({\bf z})}{q({\bf z|x})}+\log\frac{p(c|{\bf z})}{q(c|{\bf x})}\right]d{\bf z}\nonumber\\
=&\int_{\bf z}q({\bf z|x})\log\frac{p({\bf x|z})p({\bf z})}{q({\bf z|x})}d{\bf z}
-\int_{\bf z}q({\bf z|x})D_{KL}(q(c|{\bf x})||p(c|{\bf z}))d{\bf z}\label{eq:q_x_prove}
\end{flalign}
\end{small}

In Equation~\ref{eq:q_x_prove}, the first term has no relationship with $c$ and the second term is non-negative. 
Hence, to maximize $\mathcal{L}_{\textup{ELBO}}({\bf x})$, $D_{KL}(q(c|{\bf x})||p(c|{\bf z})) \equiv 0$
should be satisfied. As a result, we use the following equation to compute $q(c|{\bf x})$ in \vade:
\begin{equation}
q(c|{\bf x})=p(c|{\bf z})\equiv\frac{p(c)p({\bf z}|c)}{\sum_{c'=1}^Kp(c')p({\bf z}|c')}
\label{eqn:p_c_z}
\end{equation}

\begin{figure}[ht]
\begin{center}
\includegraphics[width=0.95\linewidth]{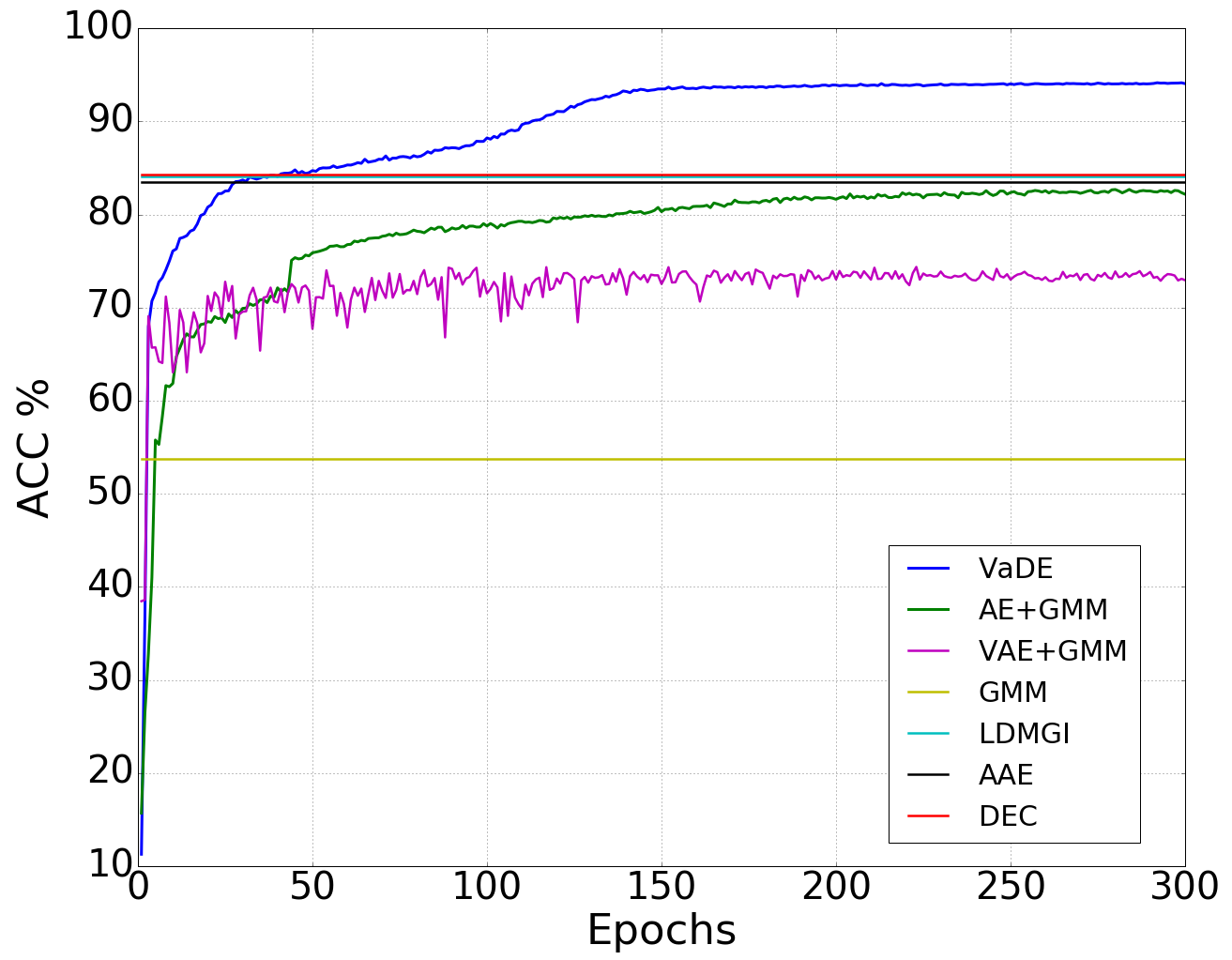}
\end{center}
\caption{Clustering accuracy over number of epochs during training on MNIST. 
We also illustrate the best performances of DEC, AAE, LDMGI and GMM.
It is better to view the figure in color.}
\label{fig:KL}
\end{figure}

By using Equation~\ref{eqn:p_c_z}, the information loss induced by the mean-field 
approximation can be mitigated, since $p(c|{\bf z})$ captures the 
relationship between $c$ and $\bf z$. It is worth noting that $p(c|{\bf z})$
is only an approximation to $q(c|{\bf x})$,
and we find it works well in practice\footnote{We approximate $q(c|{\bf x})$ by: 
1) sampling a ${\bf z}^{(i)}\sim q({\bf z|x})$; 2) computing $q(c|{\bf x})=p(c|{\bf z}^{(i)})$ 
according to Equation~\ref{eqn:p_c_z}}.

Once the training is done by maximizing the ELBO w.r.t the parameters of 
$\lbrace \boldsymbol{\pi}, \boldsymbol{\mu}_c, \boldsymbol{\sigma}_c, 
 \boldsymbol{\theta}, \boldsymbol{\phi} \rbrace$, $c \in \lbrace 1,\cdots, K\rbrace$,
a latent representation $\bf z$ 
can be extracted for each observed sample $\bf x$ by Equation~\ref{eqn:g_mu_sigma} 
and Equation~\ref{eqn:q_z_x}, and the clustering assignments can be obtained 
by Equation~\ref{eqn:p_c_z}.

\subsection{Understanding the ELBO of \vade}
\label{sec:elbo}

This section, we provide some intuitions of the ELBO of \vade. More specifically, the ELBO in Equation~\ref{eqn:loglikelihood} can be further rewritten as:
\begin{equation}
\mathcal{L}_{\textup{ELBO}}({\bf x})=E_{q({\bf z},c|{\bf x})}[\log p({\bf x}|{\bf z})]-D_{KL}(q({\bf z},c|{\bf x})||p({\bf z},c))
\label{eqn:analysis_elbo}
\end{equation}

The first term in Equation~\ref{eqn:analysis_elbo} is the {\it reconstruction} term, which encourages 
\vade to explain the dataset well. And the second term is the Kullback-Leibler divergence from the Mixture-of-Gaussians (MoG) prior $p({\bf z},c)$ to the variational posterior $q({\bf z},c|{\bf x})$, which regularizes the latent embedding $\bf z$ to lie on a MoG manifold.

\begin{figure}[ht]
\begin{center}
\subfigure[\scriptsize{Epoch 0 (11.35$\%$)}]{\includegraphics[width = 0.3\linewidth]{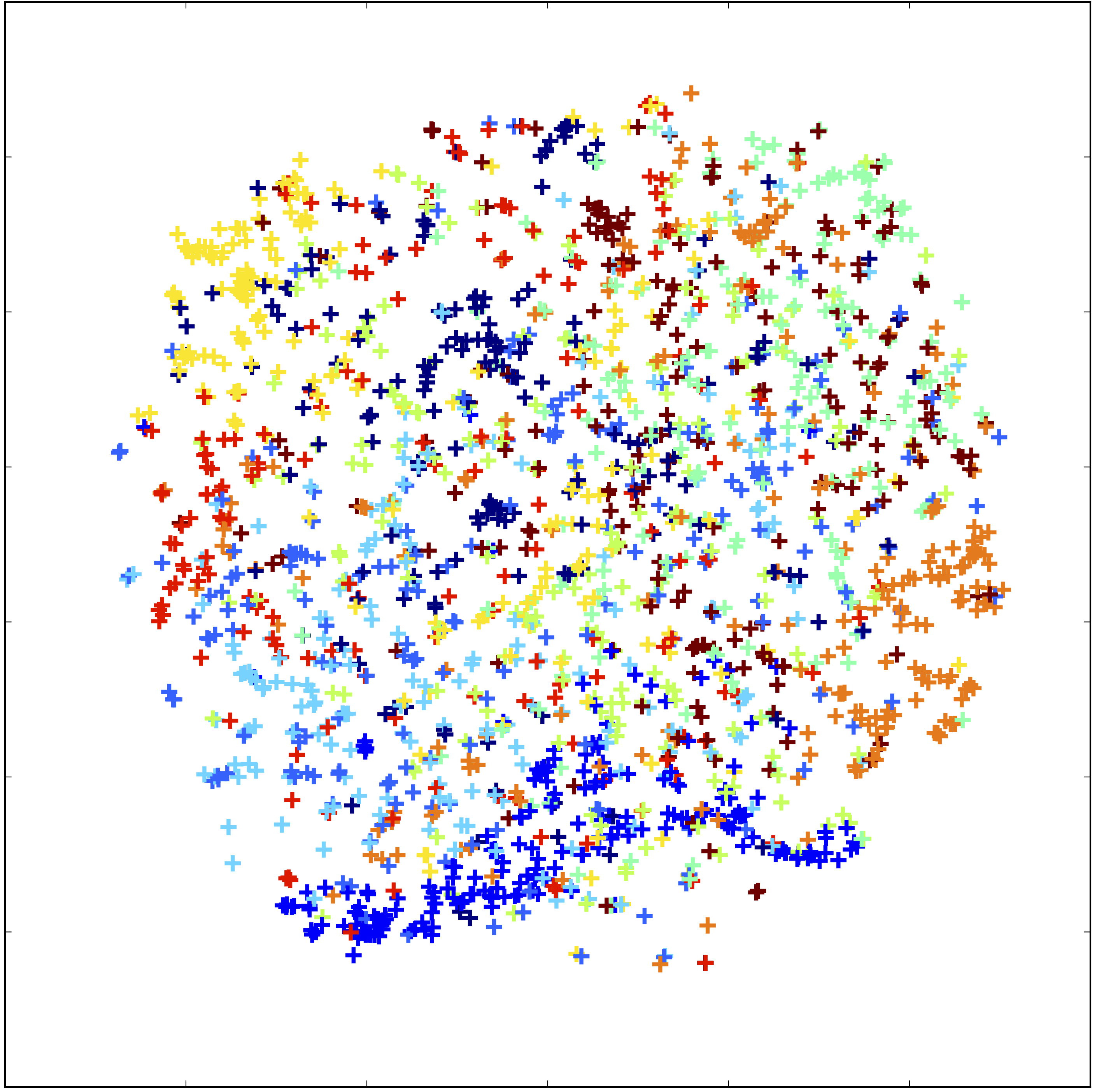}}
\subfigure[\scriptsize{Epoch 1 (55.63$\%$)}]{\includegraphics[width = 0.3\linewidth]{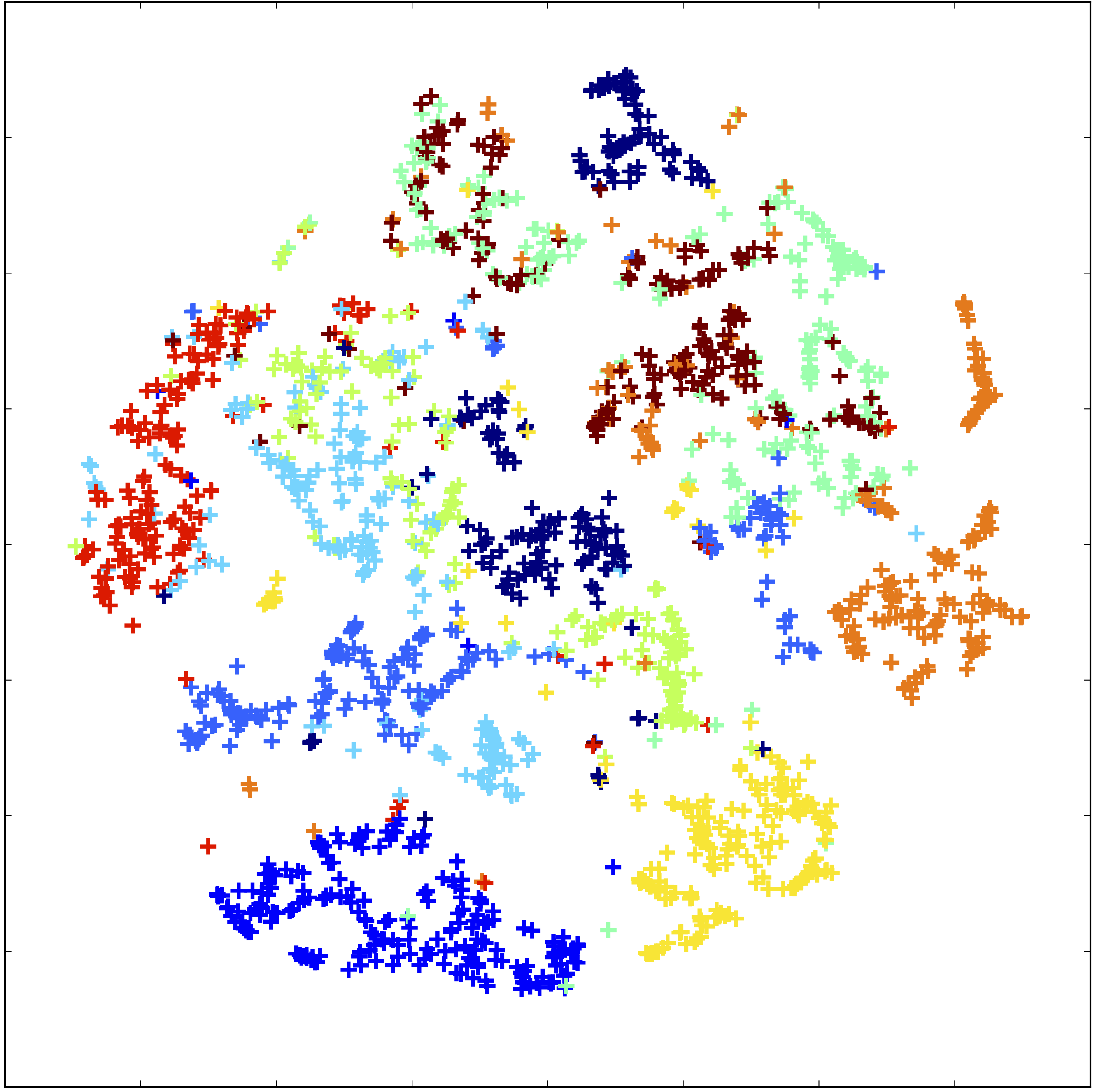}}
\subfigure[\scriptsize{Epoch 5 (72.40$\%$)}]{\includegraphics[width = 0.3\linewidth]{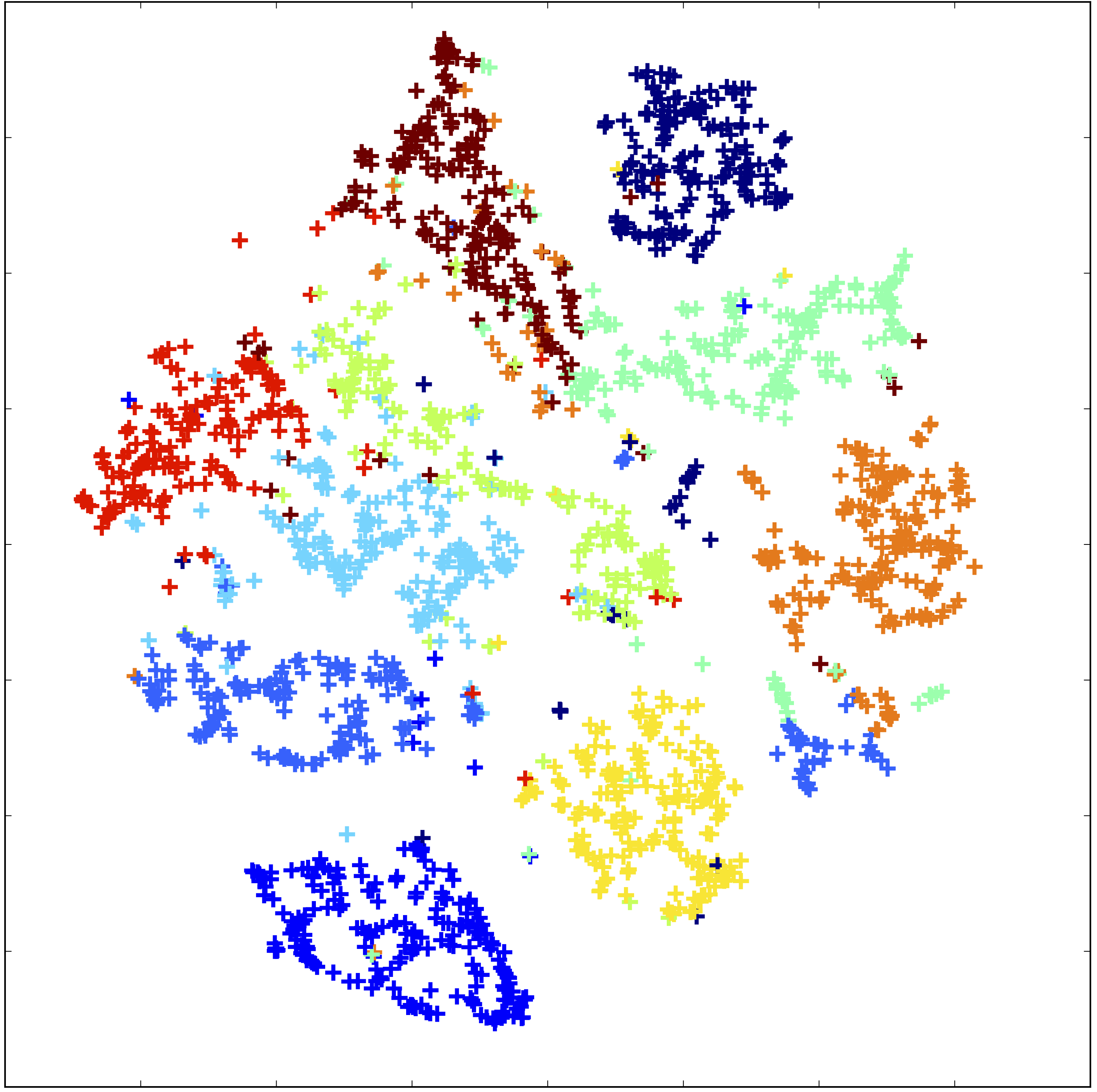}}
\subfigure[\scriptsize{Epoch 50 (84.59$\%$)}]{\includegraphics[width = 0.3\linewidth]{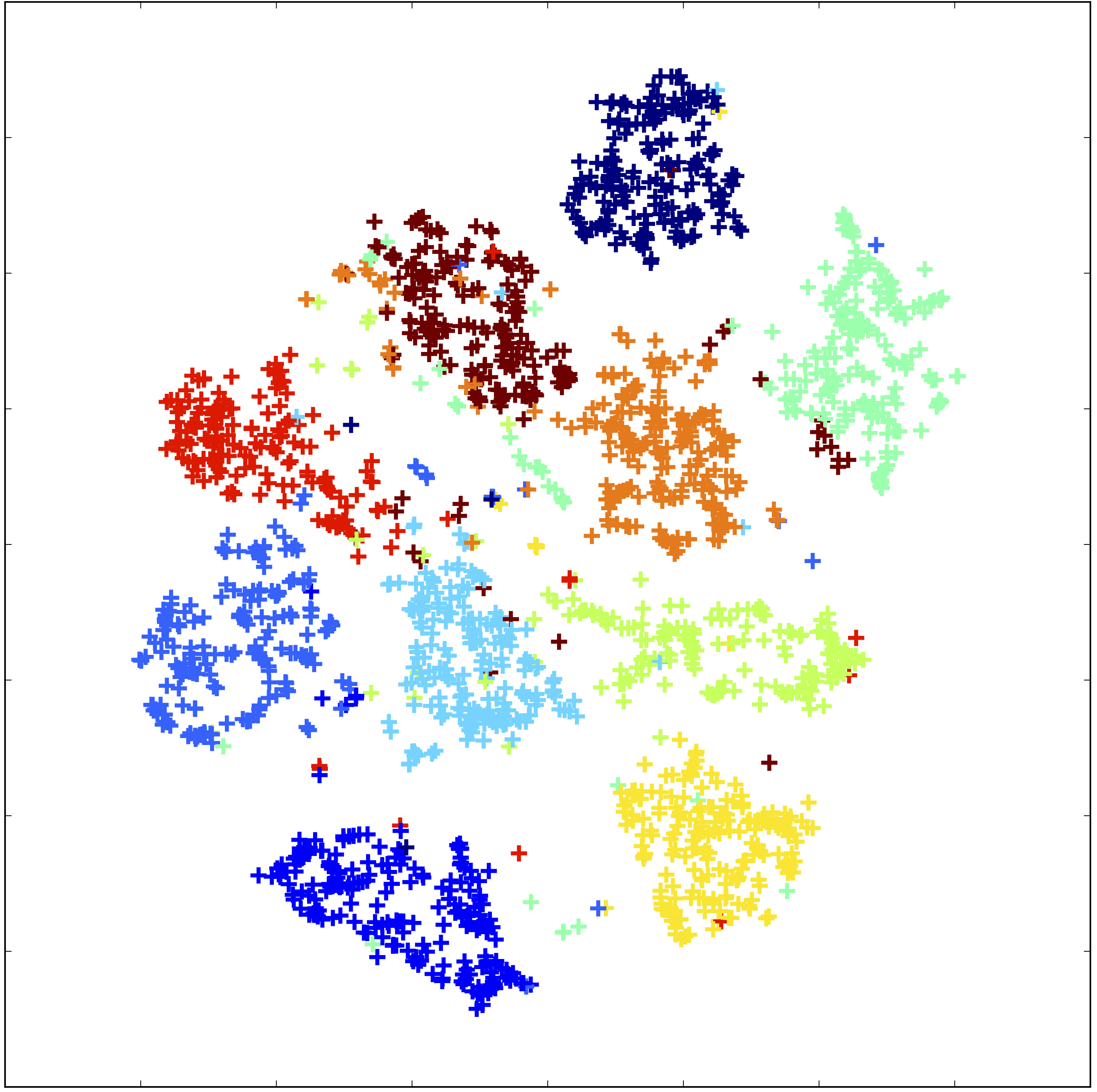}}
\subfigure[\scriptsize{Epoch 120 (90.76$\%$)}]{\includegraphics[width = 0.3\linewidth]{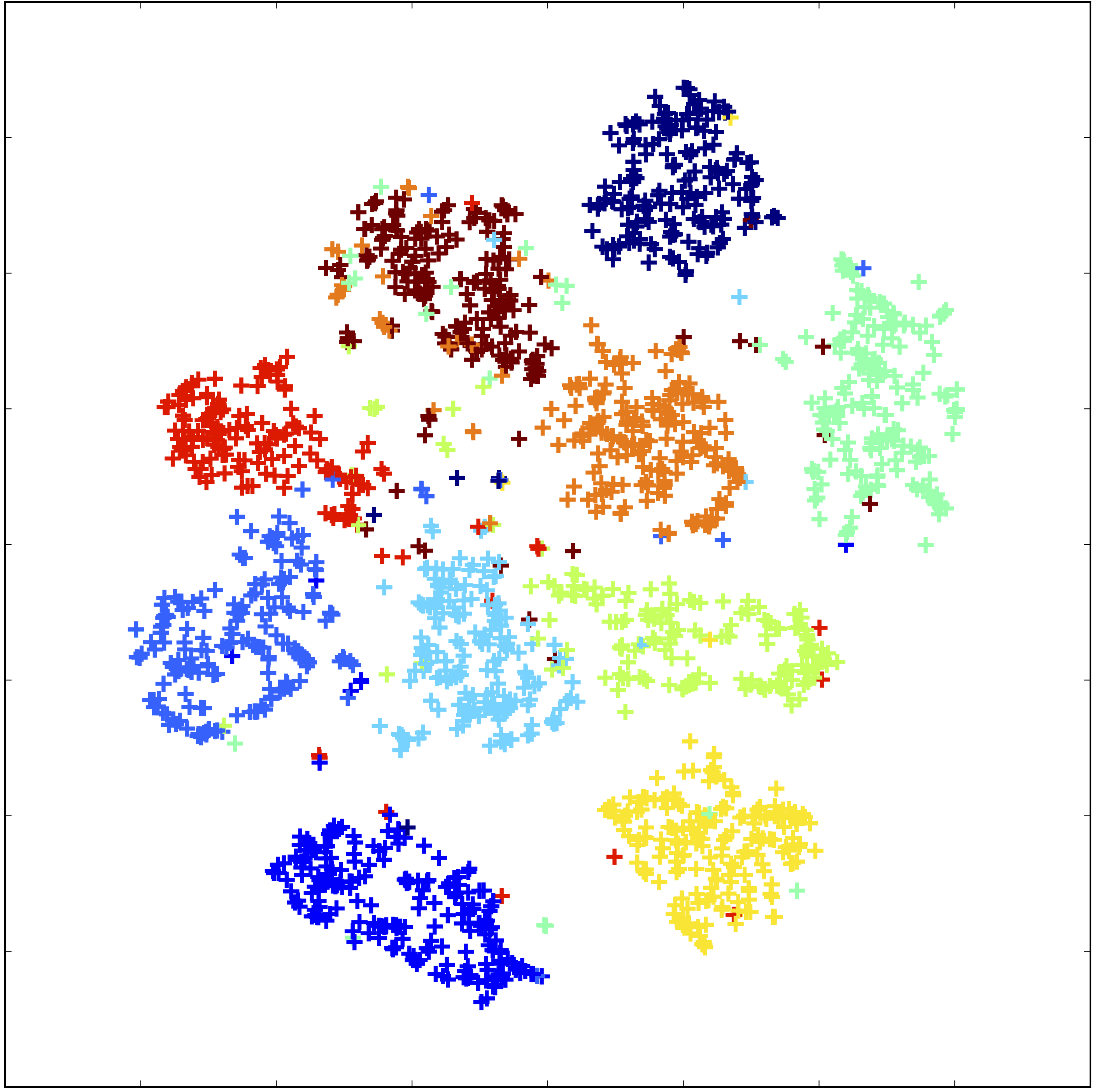}}
\subfigure[\scriptsize{Epoch End (94.46$\%$)}]{\includegraphics[width = 0.3\linewidth]{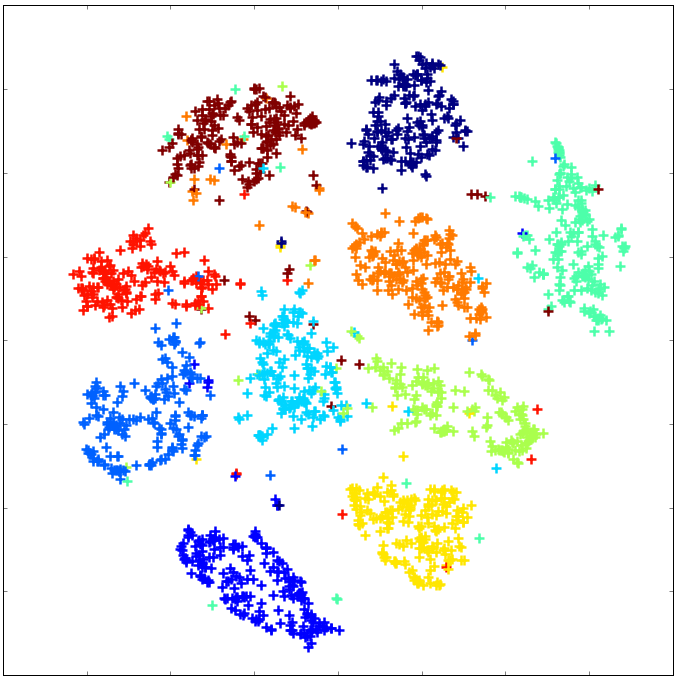}}
\end{center}
   \caption{The illustration about how data is clustered in the latent space learned by VaDE during
   training on MNIST. Different colors indicate different ground-truth classes and the clustering accuracy at the corresponding epoch is reported in the bracket. It is clear to see that the latent representations become more and more suitable for clustering during training, which can also be proved by the increasing clustering accuracy.}
\label{fig:epoch_vis}
\end{figure}

To demonstrate the importance of the KL term in Equation~\ref{eqn:analysis_elbo}, we 
train an Auto-Encoder (AE) with the same network architecture as \vade first, and then apply GMM 
on the latent representations from the learned AE, since a VaDE model without the KL term is 
almost equivalent to an AE. We refer to this model as AE+GMM.
We also show the performance of using GMM directly on the observed
space (GMM), using VAE on the observed space and 
then using GMM on the latent space from VAE (VAE+GMM)\footnote{By doing this, VAE and GMM are optimized separately.},
as well as the performances of LDMGI~\cite{yang10}, AAE~\cite{makhzani16AAE} and DEC~\cite{xie15}, in Figure~\ref{fig:KL}.
The fact that \vade outperforms AE+GMM (without KL term) and VAE+GMM significantly confirms the importance of the regularization term and the advantage of jointly optimizing VAE and GMM by \vade. We also present the illustrations of
clusters and the way they are changed w.r.t. training epochs on MNIST dataset in Figure~\ref{fig:epoch_vis}, where we map
the latent representations ${\bf z}$ into 2D space by t-SNE~\cite{maaten08}.

\section{Experiments}
\label{sec:experimens}
In this section, we evaluate the performance of \vade on 5 benchmarks from different modalities: MNIST~\cite{lecun98}, 
HHAR~\cite{stisen15}, Reuters-10K~\cite{lewis04}, Reuters~\cite{lewis04} and STL-10~\cite{coates11}. We provide quantitative comparisons of \vade with other clustering methods including GMM,
AE+GMM, VAE+GMM, LDGMI~\cite{yang10}, AAE~\cite{makhzani16AAE} and the strong baseline DEC~\cite{xie15}.
We use the 
same network architecture as DEC for a fair comparison.
The experimental results show that \vade achieves 
the state-of-the-art performance on all these benchmarks.
Additionally, we also provide quantitatively comparisons with other variants of VAE
on the discriminative quality of the latent representations.
The code of \vade is available at \url{https://github.com/slim1017/VaDE}.

\subsection{Datasets Description}
\label{sec:datasets}

The following datasets are used in our empirical experiments.
\begin{itemize}
\item {\bf MNIST}: The MNIST dataset consists of $70000$ handwritten digits. 
The images are centered and of size 28 by 28 pixels. 
We reshaped each image to a 784-dimensional vector.
\item {\bf HHAR}: The Heterogeneity Human Activity Recognition (HHAR) dataset 
contains $10299$ sensor records from smart phones and smart watches. 
All samples are partitioned into $6$ categories of human activities 
and each sample is of $561$ dimensions.
\item {\bf REUTERS}: There are around $810000$ English news stories labeled with a category tree in original Reuters dataset. Following DEC, we used $4$ root categories: corporate/industrial, government/social, markets, and economics as labels and discarded all documents with multiple labels, which results in a $685071$-article dataset.
We computed tf-idf features on the $2000$ most frequent words to represent all articles. Similar to 
DEC, a random subset of $10000$ documents is sampled, which is referred to as Reuters-10K,
since some spectral clustering methods (e.g. LDMGI)
cannot scale to full Reuters dataset.
\item {\bf STL-10}: The STL-10 dataset consists of color 
images of 96-by-96 pixel size. There are $10$ classes with $1300$ examples each.
Since clustering directly from raw pixels of high resolution images is rather difficult,
we extracted features of images of STL-10 by ResNet-50~\cite{he16}, which were then used to
test the performance of \vade and all baselines. More specifically, we applied a $3\times 3$
average pooling over the last feature map of ResNet-50 and the dimensionality 
of the features is $2048$.
\end{itemize}

\begin{table}
\begin{center}
\begin{tabular}{|l|c|c|c|}
\hline
Dataset & $\#$ Samples & Input Dim & $\#$ Clusters \\
\hline\hline
MNIST    & $70000$&$784$&$10$\\
HHAR    & $10299$&$561$&$6$\\
REUTERS-10K  & $10000$&$2000$&$4$\\
REUTERS  &$685071$&$2000$&$4$\\
STL-10  &$13000$&$2048$&$10$\\
\hline
\end{tabular}
\end{center}
\caption{Datasets statistics}
\label{table:dataset}
\end{table}

\begin{table*}
\begin{center}
\begin{small}
\begin{tabular}{|l|c|c|c|c|c|}
\hline
Method & MNIST & HHAR& REUTERS-10K & REUTERS & STL-10\\
\hline\hline
GMM   &$53.73$&$60.34$&$54.72$&$55.81$&$72.44$\\
AE+GMM & $82.18$&$77.67$&$70.13$&$70.98$&$79.83$\\
VAE+GMM & $72.94$&$68.02$&$69.56$&$60.89$&$78.86$\\
LDMGI  &$84.09^{\dagger}$&$63.43$&$65.62$&N/A&$79.22$\\
AAE & $83.48$&$83.77$&$69.82$&$75.12$&$80.01$\\
DEC  & $84.30^{\dagger}$&$79.86$&$74.32$&$75.63^{\dagger}$&$80.62$\\
VaDE  &${\bf 94.46}$&${\bf 84.46}$&${\bf 79.83}$&${\bf 79.38}$&${\bf 84.45}$\\
\hline
\end{tabular}
\end{small}
\begin{minipage}{0.5\textwidth}
{\small
$\dagger$: Taken from \cite{xie15}.
}
\end{minipage}
\end{center}
\caption{Clustering accuracy ($\%$) performance comparison on all datasets.}
\label{table:results}
\end{table*}

\begin{table}
\begin{center}
\begin{tabular}{|l|c|c|c|}
\hline
Method & k=3 & k=5 & k=10 \\
\hline\hline
VAE & $18.43$ & $15.69$ & $14.19$\\
DLGMM & $9.14$ & $8.38$ & $8.42$\\
SB-VAE & $7.64$ & $7.25$ & $7.31$\\
VaDE &${\bf 2.20}$ & ${\bf 2.14}$ & ${\bf 2.22}$\\
\hline
\end{tabular}
\end{center}
\caption{MNIST test error-rate ($\%$) for kNN on latent space.}
\label{table:sbvae}
\end{table}

\subsection{Experimental Setup}
\label{sec:exp_config}

As mentioned before, the same network architecture as DEC is adopted by \vade for a fair comparison. 
Specifically, the architectures of $f$ and $g$ in Equation~\ref{eqn:f1} and Equation~\ref{eqn:g_mu_sigma}
are $10$-$2000$-$500$-$500$-$D$ and $D$-$500$-$500$-$2000$-$10$, respectively, where $D$ is the input dimensionality.
All layers are fully connected. Adam optimizer~\cite{kingma15adam} is used to maximize the ELBO of Equation~\ref{eqn:elbo_fact},
and the mini-batch size is $100$. The learning rate for MNIST, HHAR, Reuters-10K and STL-10 is $0.002$ and 
decreases every $10$ epochs with a decay rate of $0.9$, and the learning rate for Reuters is $0.0005$ with
a decay rate of $0.5$ for every epoch.
As for the generative process in Section~\ref{sec:gen_process}, the multivariate Bernoulli distribution
is used for MNIST dataset, and the multivariate Gaussian distribution is used for the others. The number of 
clusters is fixed to the number of classes for each dataset, similar to DEC. We will 
vary the number of clusters in Section~\ref{exp:n_clusters}.

Similar to other VAE-based models~\cite{sonderby16,kingma16improving}, \vade suffers from the 
problem that the reconstruction term in Equation~\ref{eqn:analysis_elbo} would be so weak in the beginning
of training that the model might get stuck in an undesirable local minima or saddle point, from which
it is hard to escape. In this work, pretraining is used to avoid this problem. 
Specifically, we use a Stacked Auto-Encoder to pretrain the networks $f$ and $g$. 
Then all data points are projected into the latent space $\bf z$ by the pretrained network $g$, where a 
GMM is applied to initialize the parameters of 
$\lbrace \boldsymbol{\pi}, \boldsymbol{\mu}_c, \boldsymbol{\sigma}_c\rbrace$, $c \in \lbrace 1,\cdots, K\rbrace$.
In practice, few epochs of pretraining are enough to provide a good initialization of \vade. 
We find that \vade is not sensitive to hyperparameters after pretraining. Hence, we did not spend
a lot of effort to tune them. 

\subsection{Quantitative Comparison}
\label{sec:quan_comparison}

Following DEC, the performance of \vade is measured by {\it unsupervised clustering accuracy (ACC)}, which is
defined as:
\begin{equation}
\textup{ACC}=\max_{m\in \mathcal{M}}\frac{\sum_{i=1}^N{\mathds{1}}\{l_i=m(c_i)\}}{N}\nonumber
\end{equation}
where $N$ is the total number of samples, $l_i$ is the ground-truth 
label, $c_i$ is the cluster assignment obtained by the model,
and $\mathcal{M}$ is the set of all possible one-to-one mappings between cluster assignments and labels.
The best mapping can be obtained by using the Kuhn–Munkres algorithm~\cite{munkres57}. 
Similar to DEC, we perform $10$ random restarts when initializing all clustering models
and pick the result with the best objective value. As for LDMGI, AAE and DEC, we 
use the same configurations as their original papers. Table~\ref{table:results} 
compares the performance of \vade with other baselines over all datasets.
It can be seen that \vade outperforms all these baselines by a large margin on all datasets.
Specifically, on MNIST, HHAR, Reuters-10K, Reuters and STL-10 dataset, \vade achieves ACC of $94.46\%$,
$84.46\%$, $79.83\%$, $79.38\%$ and $84.45\%$, which outperforms DEC with a
relative increase ratio of $12.05\%$, $5.76\%$, $7.41\%$, $4.96\%$ and $4.75\%$, respectively.

We also compare \vade with SB-VAE~\cite{Nalisnick16SBVAE} and DLGMM~\cite{nalisnickapproximate} 
on the discriminative power of the latent representations, since these two baselines cannot
do clustering tasks. Following SB-VAE, the discriminative powers of 
the models' latent representations are assessed by running a k-Nearest Neighbors classifier (kNN)
on the latent representations of MNIST. Table~\ref{table:sbvae} shows the error rate of the kNN classifier
on the latent representations. It can be seen that \vade outperforms SB-VAE and DLGMM significantly\footnote{We use the same network architecture for \vade, SB-VAE in Table~\ref{table:sbvae}
for fair comparisons. Since there is no code available for DLGMM, we take the number of DLGMM directly
from \cite{nalisnickapproximate}. Note that \cite{Nalisnick16SBVAE} has already shown 
that the performance of SB-VAE is comparable to
DLGMM.}.

\begin{figure}[h]
\begin{center}
\subfigure[GMM]{\includegraphics[width = 0.35\linewidth]{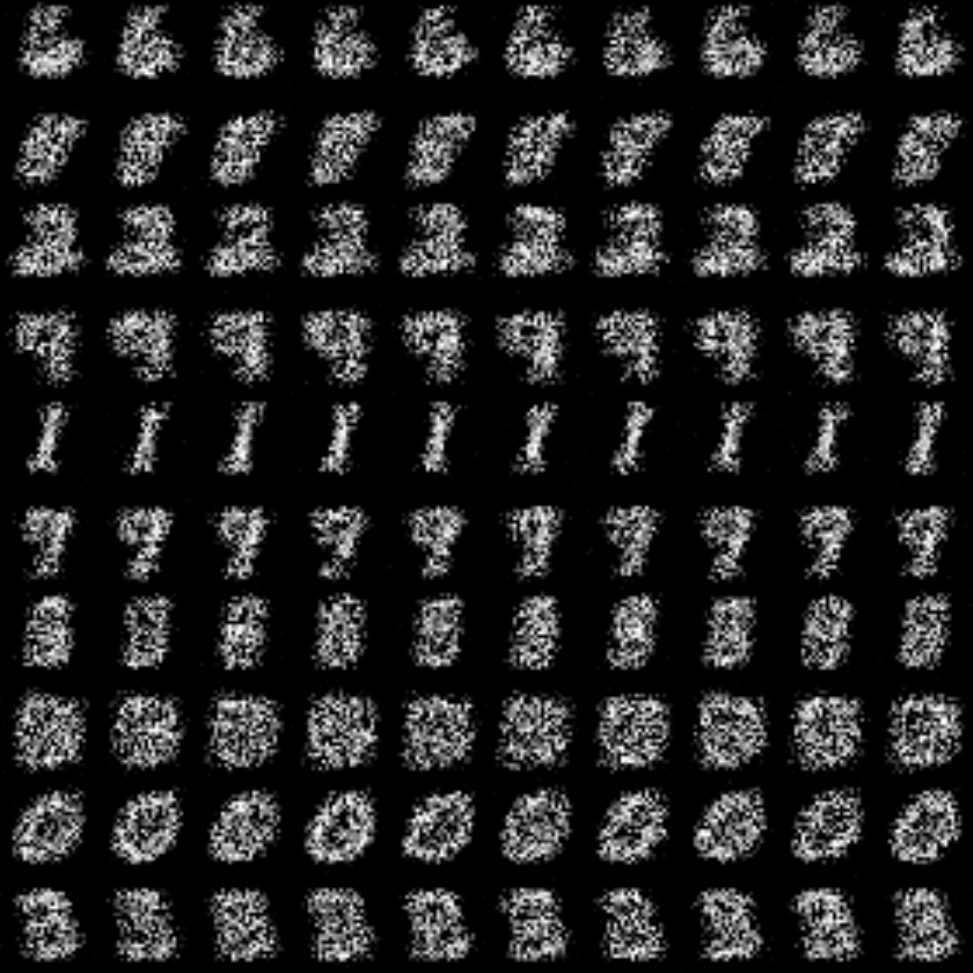}}
\hspace{6mm}
\subfigure[VAE]{\includegraphics[width = 0.35\linewidth]{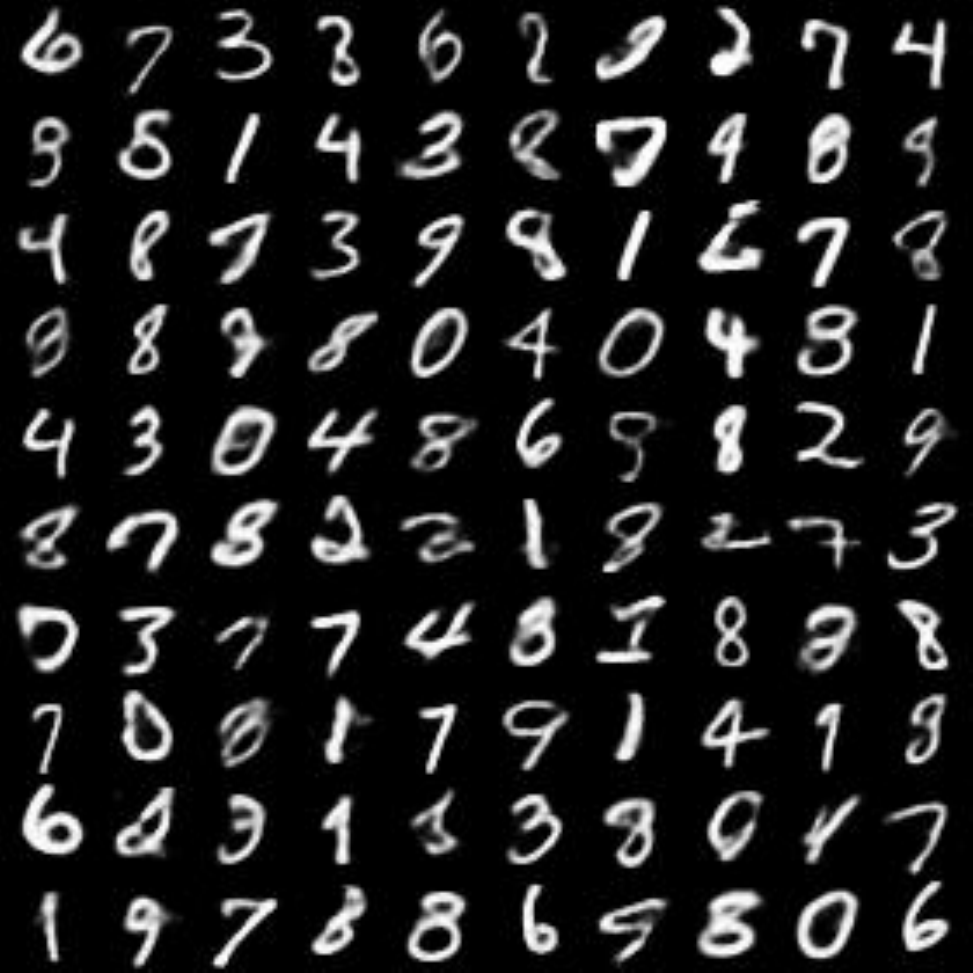}}
\subfigure[InfoGAN]{\includegraphics[width = 0.35\linewidth]{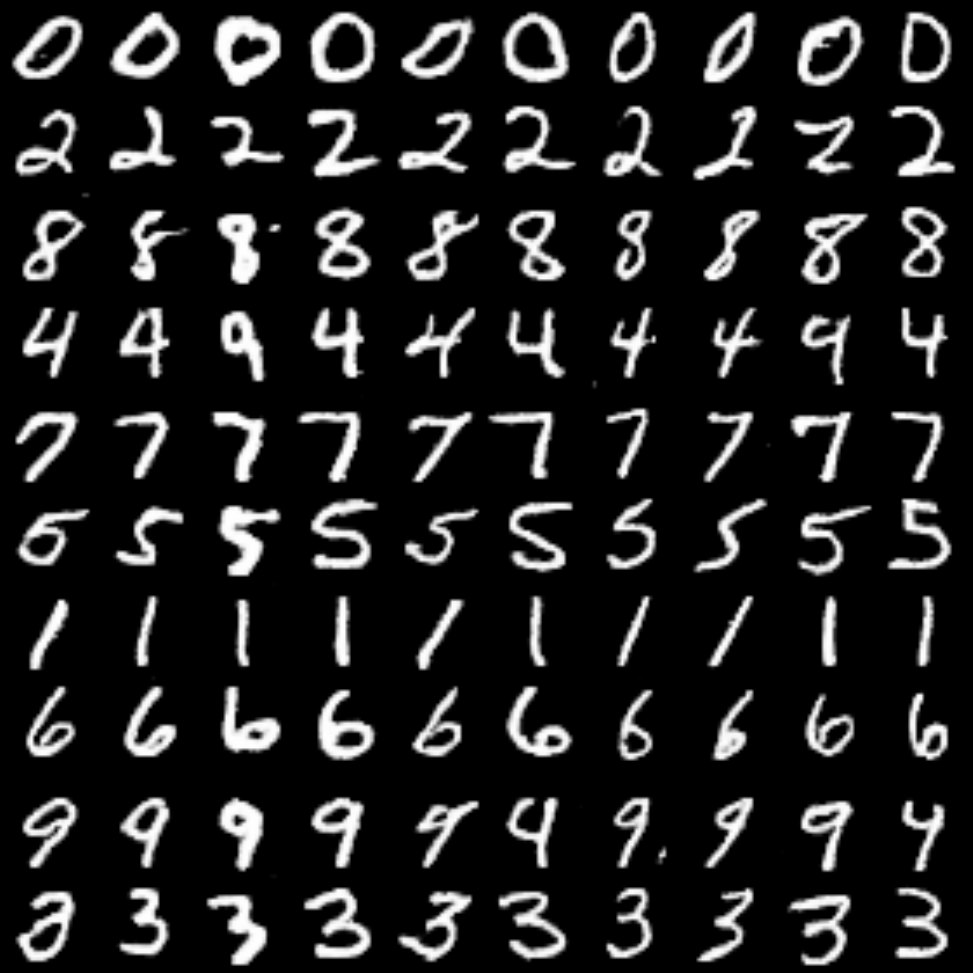}}
\hspace{6mm}
\subfigure[VaDE]{\includegraphics[width = 0.35\linewidth]{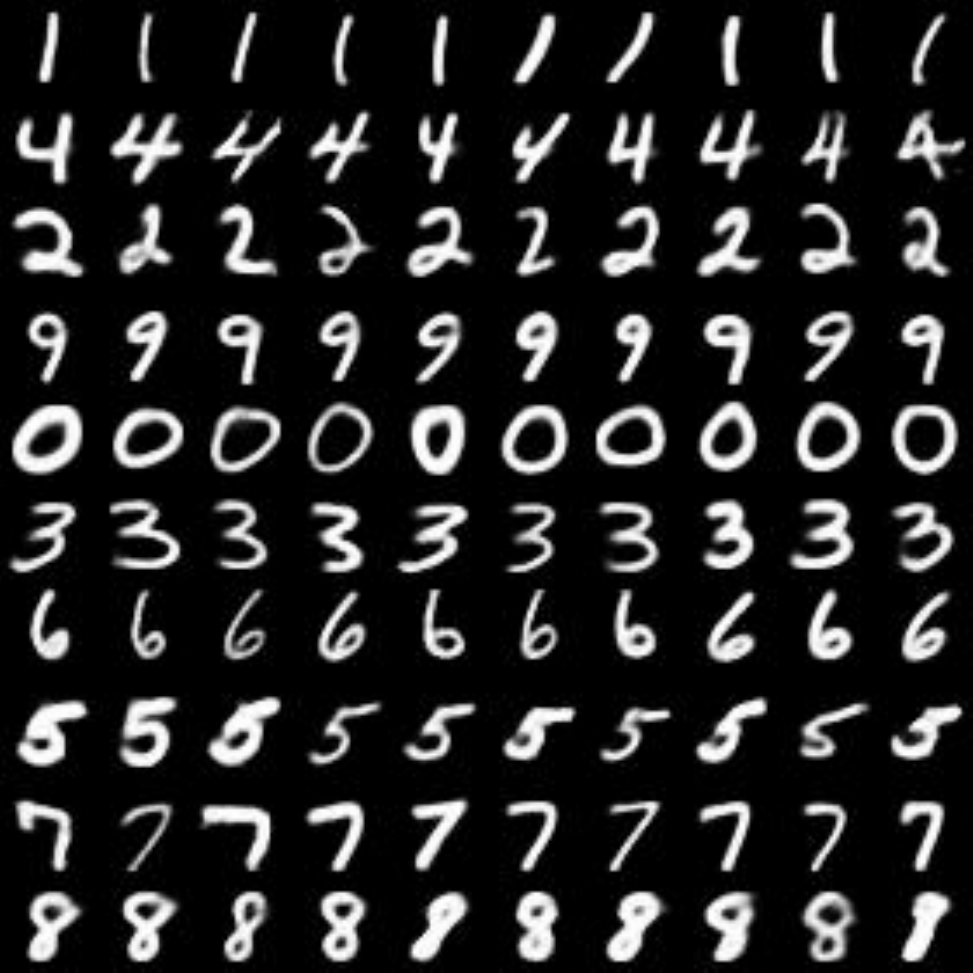}}
\end{center}
   \caption{The digits generated by GMM, VAE, InfoGAN and \vade. Except (b), digits in the same row come from the same cluster.}
\label{fig:generating}
\end{figure}

Note that although \vade can learn discriminative representations of samples, 
the training of \vade is in a totally \textit{unsupervised} way. Hence, we did not compare \vade
with other supervised models.

\subsection{Generating Samples by \vade}
\label{sec:exp_generating}
One major advantage of \vade over DEC~\cite{xie15} is that it is by nature a 
{\it generative} clustering model and can generate highly realistic samples for any specified cluster (class).
In this section, we provide some qualitative comparisons on generating samples among \vade, GMM, VAE and the state-of-art generative method InfoGAN~\cite{Chen16InfoGAN}.

Figure~\ref{fig:generating} illustrates the generated samples for class $0$ to $9$ of
MNIST by GMM, VAE, InfoGAN and \vade, respectively. It can be seen that the digits 
generated by \vade are smooth and diverse. Note that the classes of the samples from VAE
cannot be specified. We can also see that the 
performance of \vade is comparable with InfoGAN.

\subsection{Visualization of Learned Embeddings}
\label{sec:exp_visualization}

\begin{figure}[h]
\begin{center}
\includegraphics[width=1\linewidth]{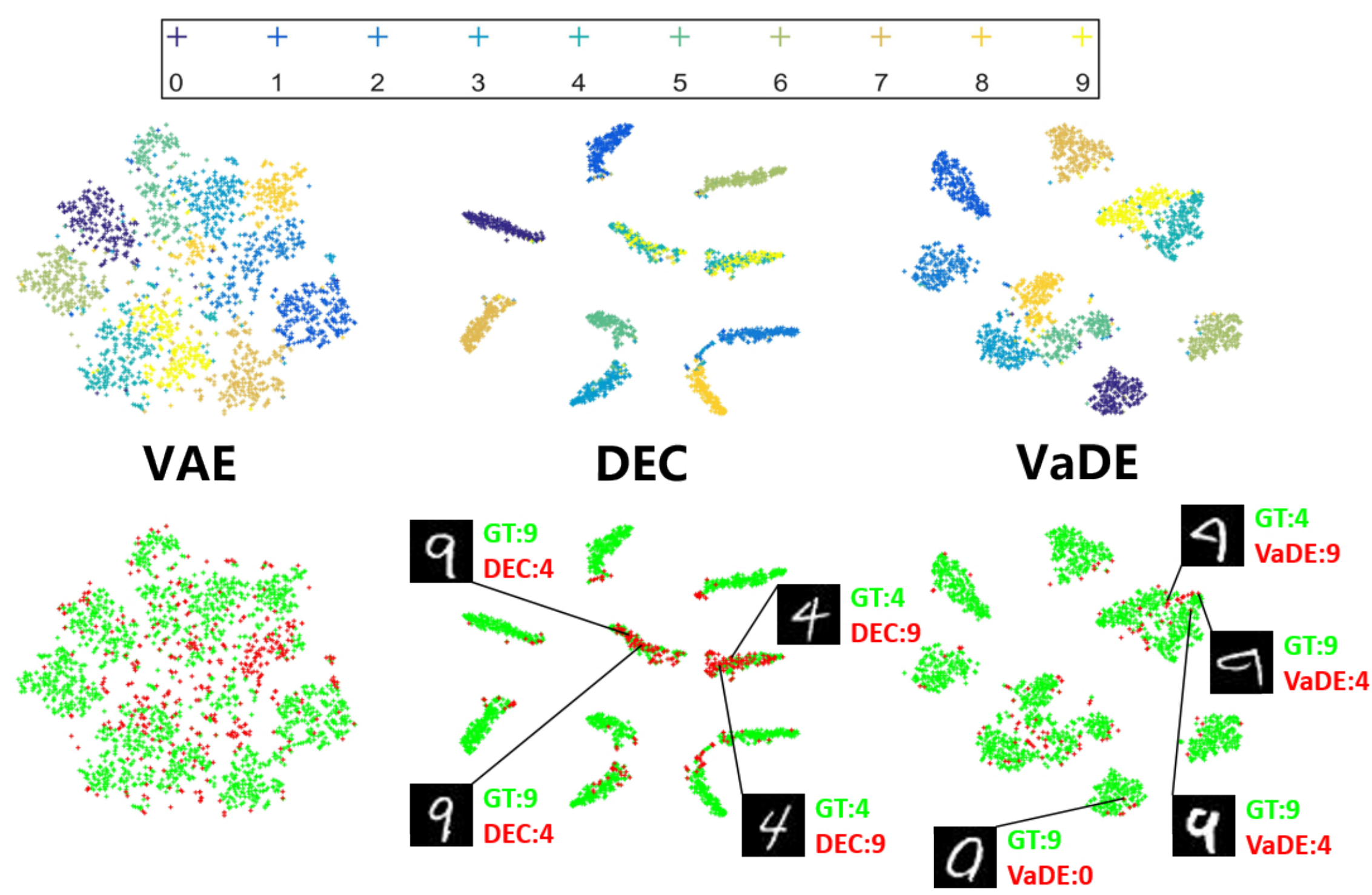}
\end{center}
   \caption{Visualization of the embeddings learned by VAE, DEC and VaDE on
   MNIST, respectively. The first row illustrates the ground-truth labels for each digit, 
   where different colors indicate different labels. The second row demonstrates the 
   clustering results, where correctly clustered samples are colored with green 
   and, incorrect ones with red. GT:4 means the ground-truth label of 
   the digit is $4$, DEC:4 means DEC assigns the digit to the cluster of 4,
   and \vade:4 denotes the assignment by \vade is $4$, and so on. 
   It is better to view the figure in color.}\label{fig:embeddings}
\end{figure}

In this section, we visualize the learned representations of VAE, 
DEC and \vade on MNIST dataset. To this end, we use t-SNE~\cite{maaten08} 
to reduce the dimensionality of the latent representation $\bf z$ from
$10$ to $2$, and plot $2000$ randomly sampled digits in Figure~\ref{fig:embeddings}. 
The first row of Figure~\ref{fig:embeddings} illustrates the ground-truth labels
for each digit, where different colors indicate different labels. The 
second row of Figure~\ref{fig:embeddings} demonstrates the clustering results, 
where correctly clustered samples are colored with green and incorrect ones with red. 

From Figure~\ref{fig:embeddings} we can see that the original VAE which
used a single Gaussian prior does not perform well in clustering tasks. 
It can also be observed that the embeddings 
learned by \vade are better than those by VAE and DEC, since the number of 
incorrectly clustered samples is smaller. Furthermore, incorrectly clustered
samples by \vade are mostly located at the border of each cluster, where 
confusing samples usually appear. In contrast, a lot of the incorrectly 
clustered samples of DEC appear in the interior of the clusters, which
indicates that DEC fails to preserve the inherent structure of the data.
Some mistakes made by DEC and \vade are also marked in Figure~\ref{fig:embeddings}.

\subsection{The Impact of the Number of Clusters}
\label{exp:n_clusters}

\begin{figure}[h]
\begin{center}
\subfigure[7 clusters]{\includegraphics[height = 0.33\linewidth]{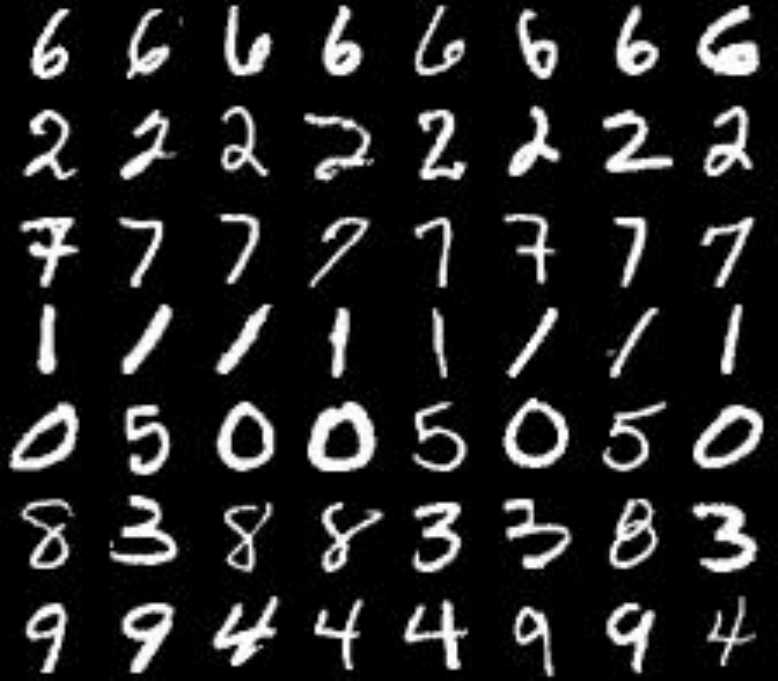}\label{fig:7_clusters}}
\hspace{10mm}
\subfigure[14 clusters]{\includegraphics[height = 0.33\linewidth]{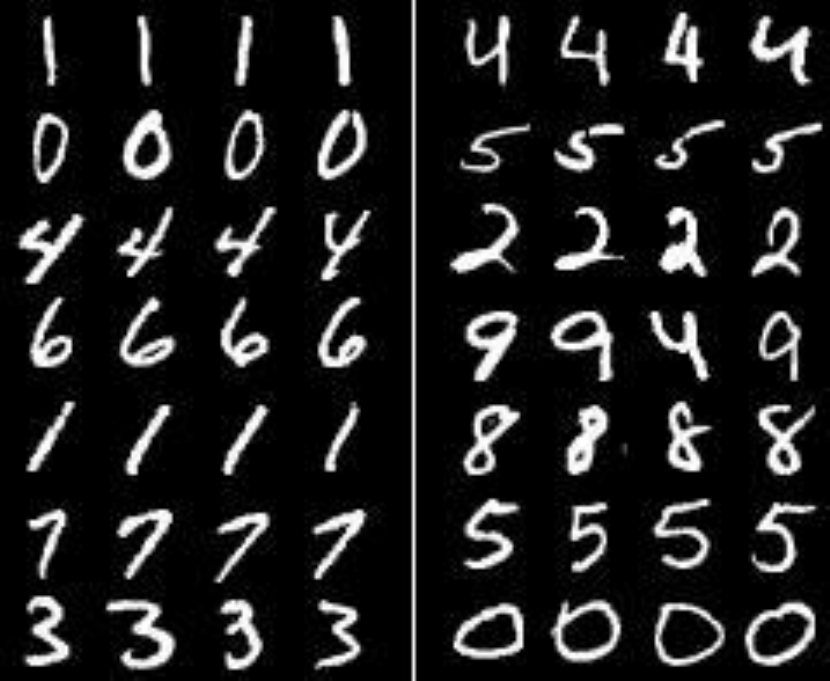}\label{fig:14_clusters}}
\end{center}
   \caption{Clustering MNIST with different numbers of clusters.
   We illustrate samples belonging to 
   each cluster by rows.}\label{fig:n_clusters}
\end{figure}

So far, the number of clusters for \vade is set to the number of classes for each dataset, which is a prior knowledge. To demonstrate \vade's representation power as an unsupervised clustering model, we deliberately choose different numbers of clusters $K$. Each row in Figure~\ref{fig:n_clusters} illustrates the samples from a cluster grouped by \vade on MNIST dataset, where $K$ is set to $7$ and $14$ in Figure~\ref{fig:7_clusters} and Figure~\ref{fig:14_clusters}, respectively. We can see that, if $K$ is smaller than the number of classes,
digits with similar appearances will be clustered together, such as $9$ and $4$, $3$ and $8$ in Figure~\ref{fig:7_clusters}.
On the other hand, if $K$ is larger than the number of classes, some digits will fall into sub-classes by \vade, such as the fatter $0$ and thinner $0$, and the upright $1$ and oblique $1$ in Figure~\ref{fig:14_clusters}.

\section{Conclusion}
\label{sec:conclusion}

In this paper, we proposed Variational Deep Embedding (\vade) which embeds the
probabilistic clustering problems into a Variational Auto-Encoder (VAE) framework. 
\vade models the data generative procedure by a GMM model and a neural network, and
is optimized by maximizing the evidence lower bound (ELBO) of the log-likelihood of
data by the SGVB estimator and the {\it reparameterization} trick. 
We compared the clustering performance of \vade with strong baselines on 5 benchmarks 
from different modalities, and the experimental results showed that \vade outperforms
the state-of-the-art methods by a large margin. We also showed that \vade could generate
highly realistic samples conditioned on cluster information without using any supervised 
information during training. Note that although we use a MoG prior for \vade in 
this paper, other mixture models can also be adopted in this framework flexibly, 
which will be our future work. 

\section*{Acknowledgments}

We thank the School of Mechanical Engineering of BIT~(Beijing Institute of Technology) and Collaborative Innovation Center of Electric Vehicles in Beijing for their support. This work was supported by the National Natural Science Foundation of China~(61620106002, 61271376). We also thank the anonymous reviewers.

\bibliographystyle{named}
\bibliography{ijcai17}

\section*{Appendix A}

In this section, we provide the derivation of $q(c|{\bf x})=E_{q({\bf z| x})}\left[p( c|{\bf z})\right]$.

The evidence lower bound $\mathcal{L}_{\textup{ELBO}}({\bf x})$ can be rewritten as:

\begin{flalign}
\mathcal{L}_{\textup{ELBO}}({\bf x})&=E_{q({\bf z},c|{\bf x})}\left[\log \frac{p({\bf x,z},c)}{q({\bf z},c|{\bf x})}\right]\nonumber\\
&=\int_{\bf z}\sum_c q({\bf z},c| {\bf x})\log\frac{p({\bf x|z})p({\bf z}|c)p({\bf} c)}{q({\bf z},c| {\bf x})}d{\bf z}\nonumber\\
&=\int_{\bf z}\sum_c q(c| {\bf x})q({\bf z|x})\log\frac{p({\bf x|z})p(c|{\bf z})p({\bf z})}{q(c|{\bf x})q({\bf z|x})}d{\bf z}\nonumber\\
&=\int_{\bf z}\sum_c q(c| {\bf x})q({\bf z|x})\left[\log\frac{p({\bf x|z})p({\bf z})}{q({\bf z|x})}+\log\frac{p(c|{\bf z})}{q(c|{\bf x})}\right]d{\bf z}\nonumber\\
&=\int_{\bf z}q({\bf z|x})\log\frac{p({\bf x|z})p({\bf z})}{q({\bf z|x})}d{\bf z}-\int_{\bf z}q({\bf z|x})\sum_cq(c|{\bf x})\log\frac{q(c|{\bf x})}{p( c|{\bf z})}d{\bf z}\nonumber\\
&=\int_{\bf z}q({\bf z|x})\log\frac{p({\bf x|z})p({\bf z})}{q({\bf z|x})}d{\bf z}-\int_{\bf z}q({\bf z|x})D_{KL}(q(c|{\bf x})||p(c|{\bf z}))d{\bf z}\label{eq:A1}
\end{flalign}

In Equation~\ref{eq:A1}, the first term does not depend on $c$ and the second term is non-negative. Thus, maximizing the lower bound $\mathcal{L}_{\textup{ELBO}}({\bf x})$ with respect to $q(c|{\bf x})$ requires that $D_{KL}(q(c|{\bf x})||p(c|{\bf z}))=0$.
Thus, we have

\begin{equation}
\frac{q( c| {\bf x})}{p(c|{\bf z})}=\nu\nonumber
\end{equation} 
where $\nu$ is a constant.

Since $\sum_{c}q(c|{\bf x})=1$ and $\sum_{c}p(c|{\bf z})=1$, we have: 
\begin{equation}
\frac{q( c| {\bf x})}{p(c|{\bf z})}=1\nonumber
\end{equation}

Taking the expectation on both sides, we can obtain:
\begin{equation}
q(c|{\bf x})=E_{q({\bf z| x})}[p(c|{\bf z})]\nonumber
\end{equation}

\section*{Appendix B}
\textbf{Lemma 1} Given two multivariate Gaussian distributions $q({\bf z})=\mathcal{N}({\bf z};\boldsymbol{\tilde\mu},\boldsymbol{\tilde\sigma}^2{\bf I})$ and $p({\bf z})=\mathcal{N}({\bf z};\boldsymbol \mu,\boldsymbol{\sigma}^2{\bf I})$, we have:
\begin{flalign}
\int q({\bf z})\log p({\bf z})\,d{\bf z}
&=\sum_{j=1}^J-\frac{1}{2}\log{(2\pi \sigma_j^2)}-\frac{\tilde\sigma_j^2}{2\sigma_j^2}-\frac{(\tilde\mu_j-\mu_j)^2}{2\sigma_j^2}\label{eq:lemma}
\end{flalign}
where $\mu_j$, $\sigma_j$, ${\tilde\mu}_j$ and ${\tilde\sigma}_j$ simply denote the $j$\textsuperscript{th} element of $\boldsymbol{\mu}$, $\boldsymbol{\sigma}$, $\boldsymbol{\tilde\mu}$ and $\boldsymbol{\tilde\sigma}$, respectively, and $J$ is the dimensionality of $\bf{z}$.\vspace{5mm}

\textbf{Proof (of Lemma 1).}

\begin{flalign}
&\int q({\bf z})\log p({\bf z})\,d{\bf z}
=\int \mathcal{N}({\bf z};\boldsymbol{\tilde\mu},\boldsymbol{\tilde\sigma}^2{\bf I})\log{\mathcal{N}({\bf z};\boldsymbol \mu,\boldsymbol{\sigma}^2{\bf I})}\,d{\bf z}&\nonumber\\
=&\int\prod_{j=1}^J\frac{1}{\sqrt{2\pi\tilde\sigma_j^2}}\exp(-\frac{(z_j-\tilde\mu_j)^2}{2\tilde\sigma_j^2})\log\left[{\prod_{j=1}^J\frac{1}{\sqrt{2\pi\sigma_j^2}}\exp(-\frac{(z_j-\mu_j)^2}{2\sigma_j^2}})\right]\,d{\bf z}\nonumber\\
=&\sum_{j=1}^J\int \frac{1}{\sqrt{2\pi\tilde\sigma_j^2}}\exp(-\frac{(z_j-\tilde\mu_j)^2}{2\tilde\sigma_j^2})\log\left[{\frac{1}{\sqrt{2\pi\sigma_j^2}}\exp(-\frac{(z_j-\mu_j)^2}{2\sigma_j^2}})\right]\,d{z_j}\nonumber\\
=&\sum_{j=1}^J\int \frac{1}{\sqrt{2\pi\tilde\sigma_j^2}}\exp(-\frac{(z_j-\tilde\mu_j)^2}{2\tilde\sigma_j^2})\left[-\frac{1}{2}\log (2\pi\sigma_j^2)\right]\,d{z_j}-\int\frac{1}{\sqrt{2\pi\tilde\sigma_j^2}}\exp(-\frac{(z_j-\tilde\mu_j)^2}{2\tilde\sigma_j^2})\frac{(z_j-\mu_j)^2}{2\sigma_j^2}\,d{z_j}\nonumber\\
=&\sum_{j=1}^J -\frac{1}{2}\log (2\pi\sigma_j^2)-\int\frac{1}{\sqrt{2\pi\tilde\sigma_j^2}}\exp(-\frac{(z_j-\tilde\mu_j)^2}{2\tilde\sigma_j^2})\frac{(z_j-\tilde\mu_j)^2+2(z_j-\tilde\mu_j)(\tilde\mu_j-\mu_j)+(\tilde\mu_j-\mu_j)^2}{2\tilde\sigma_j^2}\frac{\tilde\sigma_j^2}{\sigma_j^2}\,d{z_j}\nonumber\\
=&C-\frac{\tilde\sigma_j^2}{\sigma_j^2}\int\frac{1}{\sqrt{2\pi\tilde\sigma_j^2}}\exp(-\frac{(z_j-\tilde\mu_j)^2}{2\tilde\sigma_j^2})\frac{(z_j-\tilde\mu_j)^2}{2\tilde\sigma_j^2}\,d{z_j}\nonumber
-\int\frac{1}{\sqrt{2\pi\tilde\sigma_j^2}}\exp(-\frac{(z_j-\tilde\mu_j)^2}{2\tilde\sigma_j^2})\frac{(\tilde\mu_j-\mu_j)^2}{2\sigma_j^2}\,d{z_j}&\nonumber\\
=&C-\frac{\tilde\sigma_j^2}{\sigma_j^2}\int\frac{1}{\sqrt{2\pi}}\exp(-\frac{x_j^2}{2})\frac{x_j^2}{2}\,d{x_j}-\frac{(\tilde\mu_j-\mu_j)^2}{2\sigma_j^2}\nonumber\\
=&C-\frac{\tilde\sigma_j^2}{\sigma_j^2}\int\frac{1}{\sqrt{2\pi}}(-\frac{x_j}{2})\,d{(\exp(-\frac{x_j^2}{2}))}-\frac{(\tilde\mu_j-\mu_j)^2}{2\sigma_j^2}\nonumber\\
=&C-\frac{\tilde\sigma_j^2}{\sigma_j^2}\left[\frac{1}{\sqrt{2\pi}}(-\frac{x_j}{2})\exp(-\frac{x_j^2}{2})\Big|_{-\infty}^{\infty}-\int\frac{1}{\sqrt{2\pi}}\exp(-\frac{x_j^2}{2})\,d{(-\frac{x_j}{2})}\right]-\frac{(\tilde\mu_j-\mu_j)^2}{2\sigma_j^2}\nonumber\\
=&\sum_{j=1}^J-\frac{1}{2}\log{(2\pi \sigma_j^2)}-\frac{\tilde\sigma_j^2}{2\sigma_j^2}-\frac{(\tilde\mu_j-\mu_j)^2}{2\sigma_j^2}\nonumber
\end{flalign}
where $C$ denotes $\sum_{j=1}^J -\frac{1}{2}\log (2\pi\sigma_j^2)$ for simplicity.

\section*{Appendix C}
In this section, we describe how to compute the evidence lower bound of VaDE.
Specifically, the evidence lower bound can be rewritten as:
\begin{flalign}
\mathcal{L}_{\textup{ELBO}}({\bf x})=&E_{q({\bf z},c|{\bf x})}\left[\log p({\bf x|z})\right]\nonumber\\
&+E_{q({\bf z},c|{\bf x})}\left[\log p({\bf z}|c)\right]\nonumber\\
&+E_{q({\bf z},c|{\bf x})}\left[\log p( c)\right]\nonumber\\
&-E_{q({\bf z},c|{\bf x})}\left[\log q({\bf z|x})\right]\nonumber\\
&-E_{q({\bf z},c|{\bf x})}\left[\log q(c|{\bf x})\right]\label{apeq:ELBO}
\end{flalign}
The Equation~\ref{apeq:ELBO} can be computed by substituting Equation~4, 5, 6, 11 and 16
into Equation~\ref{apeq:ELBO} and using \textbf{Lemma 1} in Appendix B. Specifically, each 
item of Equation~\ref{apeq:ELBO} can be obtained as follows:

\begin{itemize}
    \item $E_{q({\bf z},c|{\bf x})}\left[\log p({\bf x|z})\right]$:
    
    Recall that the observation {\bf x} can be modeled as either a multivariate Bernoulli distribution or a multivariate Gaussian distribution. We provide the derivation of $E_{q({\bf z},c|{\bf x})}\left[\log p({\bf x|z})\right]$ for the multivariate Bernoulli distribution, and the derivation for the multivariate Gaussian case can be obtained in a similar way. 
    
    Using the SGVB estimator, we can approximate the 
    $E_{q({\bf z},c|{\bf x})}\left[\log p({\bf x|z})\right]$ as:
    \begin{eqnarray}
    E_{q({\bf z},c|{\bf x})}\left[\log p({\bf x|z})\right]&=&\frac{1}{L}\sum_{l=1}^L\log p({\bf x}|{\bf z}^{(l)})\nonumber\\
    &=&\frac{1}{L}\sum_{l=1}^L\sum_{i=1}^D{x_i}\log{{\mu}^{(l)}_x}_i+(1-x_i)\log(1-{{\mu}^{(l)}_x}_i)\nonumber
    \end{eqnarray}
    where ${\boldsymbol{\mu}_x}^{(l)}=f({\bf z}^{(l)};{\boldsymbol{\theta}})$, ${\bf z}^{(l)}\sim \mathcal{N}(\boldsymbol{\tilde\mu},\boldsymbol{\tilde\sigma}^2{\bf I})$ and $\left[\boldsymbol{\tilde\mu}; \log \boldsymbol{\tilde\sigma}^2\right]=g({\bf x};\boldsymbol{\phi})$. $L$ is the number of Monte Carlo samples 
    in the SGVB estimator and can be set to $1$. $D$ is the dimensionality of ${\bf x}$.

    Since the Monte Carlo estimate of the expectation above is non-differentiable w.r.t $\boldsymbol{\phi}$ 
    when $z^{(l)}$ is directly sampled from ${\bf z}\sim \mathcal{N}(\boldsymbol{\tilde\mu},\boldsymbol{\tilde\sigma}^2{\bf I})$,
    we use the {\it reparameterization} trick to obtain a differentiable estimation:
    \begin{equation}
    {\bf z}^{(l)}=\boldsymbol{\tilde\mu}+\boldsymbol{\tilde\sigma}\circ\boldsymbol{\epsilon}^{(l)}\hspace{3mm}\textup{and}\hspace{3mm}\boldsymbol{\epsilon}^{(l)}\sim \mathcal{N}(0 ,{\bf I})\nonumber
    \end{equation}
    where $\circ$ denotes the element-wise product.
    
    \item $E_{q({\bf z},c|{\bf x})}\left[\log p({\bf z}|c)\right]$:
    
    \begin{eqnarray}
    E_{q({\bf z},c|{\bf x})}\left[\log p({\bf z}|c)\right]&=&\int_{\bf z}\sum_{c=1}^Kq(c| {\bf x})q({\bf z}|{\bf x})\log p({\bf z}| c)\,d{\bf z}\nonumber\\
    &=&\sum_{c=1}^Kq(c|{\bf x})\int_{\bf z}\mathcal{N}({\bf z}|\boldsymbol{\tilde\mu},\boldsymbol{\tilde\sigma}^2{\bf I})\log \mathcal{N}({\bf z}|\boldsymbol{\mu}_c,\boldsymbol{\sigma}^2_c{\bf I})\,d{\bf z}\nonumber
    \end{eqnarray}
    
    According to Lemma 1 in Appendix B, we have:
    \begin{eqnarray}
    E_{q({\bf z},c|{\bf x})}\left[\log p({\bf z}|c)\right]=-\sum_{c=1}^Kq(c|{\bf x})\left[\frac{J}{2}\log(2\pi)+\frac{1}{2}(\sum_{j=1}^J\log\sigma^2_{cj}+\sum_{j=1}^J\frac{\tilde\sigma^2_j}{\sigma^2_{cj}}+\sum_{j=1}^J\frac{(\tilde\mu_j-\mu_{cj})^2}{\sigma^2_{cj}})\right]&\label{eq:ELBO2}\nonumber
    \end{eqnarray}
    
    \item $E_{q({\bf z},c|{\bf x})}\left[\log p(c)\right]$:
    \begin{eqnarray}
    E_{q({\bf z},c|{\bf x})}[\log p(c)]&=&\int_{\bf z}\sum_{c=1}^Kq({\bf z}|{\bf x})q(c| {\bf x})\log p(c)\,d{\bf z}\nonumber\\
    &=&\int_{\bf z}q({\bf z}|{\bf x})\sum_{c=1}^Kq(c|{\bf x})\log \pi_c\,d{\bf z}\nonumber\\
    &=&\sum_{c=1}^Kq(c|{\bf x})\log \pi_c\label{eq:ELBO3}\nonumber
    \end{eqnarray}
    
    \item $E_{q({\bf z},c|{\bf x})}\left[\log q({\bf z|x})\right]$:
    \begin{eqnarray}
    E_{q({\bf z},c|{\bf x})}\left[\log q({\bf z|x})\right]&=&\int_{\bf z}\sum_{c=1}^Kq(c| {\bf x})q({\bf z}|{\bf x})\log q({\bf z}|{\bf x})\,d{\bf z}\nonumber\\
    &=&\int_{\bf z}\mathcal{N}({\bf z};\boldsymbol{\tilde\mu},\boldsymbol{\tilde\sigma}^2{\bf I})\log \mathcal{N}({\bf z};\boldsymbol{\tilde\mu},\boldsymbol{\tilde\sigma}^2{\bf I})\,d{\bf z}\nonumber
    \end{eqnarray}
    
    According to Lemma 1 in Appendix B, we have:
    \begin{eqnarray}
    E_{q({\bf z},c|{\bf x})}\left[\log q({\bf z|x})\right]=-\frac{J}{2}\log(2\pi)-\frac{1}{2}\sum_{j=1}^J(1+\log\tilde\sigma^2_j)&\label{eq:ELBO4}\nonumber
    \end{eqnarray}
    
    \item $E_{q({\bf z},c|{\bf x})}\left[\log q(c|{\bf x})\right]$:
    \begin{eqnarray}
    E_{q({\bf z},c|{\bf x})}\left[\log q(c|{\bf x})\right]&=&\int_{\bf z}\sum_{c=1}^Kq({\bf z}|{\bf x})q(c|{\bf x})\log q(c|{\bf x})\,d{\bf z}\nonumber\\
    &=&\int_{\bf z}q({\bf z}|{\bf x})\sum_{c=1}^Kq(c|{\bf x})\log q(c|{\bf x})\,d{\bf z}\nonumber\\
    &=&\sum_{c=1}^Kq(c|{\bf x})\log q(c|{\bf x})\label{eq:ELBO5}\nonumber
    \end{eqnarray}
    
\end{itemize}

\noindent where $\mu_{cj}$, $\sigma_{cj}$, ${\tilde\mu}_j$ and ${\tilde\sigma}_j$ simply denote the $j$\textsuperscript{th} element of $\boldsymbol{\mu}_c$, $\boldsymbol{\sigma}_c$, $\boldsymbol{\tilde\mu}$ and $\boldsymbol{\tilde\sigma}$ described in Section~3 respectively. $J$ is the dimensionality of {\bf z} and $K$ is the number of clusters.
    
For all the above equations, $q(c|{\bf x})$ is computed by Appendix A and can be approximated by
the SGVB estimator and the {\it reparameterization} trick as follows:
\begin{equation}
 q(c|{\bf x})=E_{q({\bf z| x})}\left[p(c|{\bf z})\right]=\frac{1}{L}\sum_{l=1}^L\frac{p(c)p({\bf z}^{(l)}|c)}{\sum_{c'=1}^Kp(c')p({\bf z}^{(l)}|c')}\label{eq:q_c_x}\nonumber
\end{equation}
where ${\bf z}^{(l)}\sim \mathcal{N}(\boldsymbol{\tilde\mu},\boldsymbol{\tilde\sigma}^2{\bf I})$, $\left[\boldsymbol{\tilde\mu}; \log \boldsymbol{\tilde\sigma}^2\right]=g({\bf x};\boldsymbol{\phi})$, ${\bf z}^{(l)}=\boldsymbol{\tilde\mu}+\boldsymbol{\tilde\sigma}\circ\boldsymbol{\epsilon}^{(l)}$ and $\boldsymbol{\epsilon}^{(l)}\sim \mathcal{N}(0 ,{\bf I})$.

\end{document}